\documentclass{tADR2earxiv}

\usepackage{url}
\usepackage{arydshln}

\usepackage{color}
\usepackage{amsmath}

\newcommand{\red}[1]{\textcolor{black}{#1}}
\newcommand{\blue}[1]{\textcolor{black}{#1}}

\begin{document}

\title{
World Robot Challenge 2020 --- Partner Robot: \\ A Data-Driven Approach for Room Tidying with Mobile Manipulator
}

\author{
T. Matsushima$^{a}$$^{\ast}$\thanks{$^\ast$Corresponding author. Email: matsushima@weblab.t.u-tokyo.ac.jp\vspace{6pt}},
Y, Noguchi$^{a}$,
J. Arima$^{b}$,
T. Aoki$^{a}$,
Y. Okita$^{a}$,
Y. Ikeda$^{a}$,
K. Ishimoto$^{a}$,
S. Taniguchi$^{a}$,
Y. Yamashita$^{a}$,
S. Seto$^{a}$,
S. Gu$^{a}$,
Y. Iwasawa$^{a}$,
and
Y. Matsuo$^{a}$
\\\vspace{6pt}
$^{a}${\em{The University of Tokyo, Tokyo, Japan}};
$^{b}${\em{Matsuo Institute, Tokyo, Japan}} 
}

\maketitle

\begin{abstract}
Tidying up a household environment using a mobile manipulator poses various challenges in robotics, such as adaptation to large real-world environmental variations, and safe and robust deployment in the presence of humans.
The Partner Robot Challenge in World Robot Challenge (WRC) 2020, a global competition held in September 2021, benchmarked tidying tasks in the real home environments, \red{and importantly, tested for \textit{full} system performances}.
For this challenge, we developed an entire household service robot system, which leverages a data-driven approach to adapt to numerous edge cases that occur during the execution, instead of classical manual pre-programmed solutions.
In this paper, we describe the core ingredients of the proposed robot system, including visual recognition, object manipulation, and motion planning.
Our robot system won the second prize, verifying the effectiveness and potential of data-driven robot systems for mobile manipulation in home environments.

\medskip

\medskip

\end{abstract}

\renewcommand{\thefootnote}{}
\footnote[0]{This is a preprint of an article submitted for consideration in ADVANCED ROBOTICS, copyright Taylor \& Francis and Robotics Society of Japan; ADVANCED ROBOTICS is available online at http://www.tandfonline.com/.}
\renewcommand{\thefootnote}{\arabic{footnote}}

\section{Introduction}
\label{sec:intro}

The World Robot Challenge 2020 (WRC2020) is a global robot competition that was held in Aichi, Japan, September 2021.
Among the challenges in the competition, the Partner Robot Challenge (Real Space) aimed to develop a robot that can support daily human activities, and approximate tidying up in real-world household environments.
Tidying up a room may seem easy for most humans (though boring for many people); however, for robots, it is a complicated task filled with difficulties.
For example, the robot needs to generalize the rich diversity of objects in real environments, adapting quickly to different types, poses, and even unseen scenarios.
Moreover, the robot needs to avoid the obstacles for safe deployment while moving both quickly and smoothly to not disrupt daily human life. 
Furthermore, unlike in a simulator, the robot cannot get access to ground-truth information about the environment and needs to decide actions promptly between each real-time control commands.

This paper describes an entire robot system that won second prize in the WRC2020 competition. 
In particular, we describe how we leveraged a data-driven approach to handle the variety of household environments. 
In real environments, any two homes do not have the same objects, furniture, or layouts.  
Therefore, it is almost impossible for the developer to enumerate all possible situations and manually implement each desired robot behavior. 
In other words, the robot should be able to generalize or adapt to the current environment rather than strictly following the pre-defined program. 
Accordingly, we utilized a data-driven approach on both perception and control of the robot system so that the robot could determine its behavior in new environments. 
For instance, we utilize and evaluate the following data-driven modules on the household environment:

\begin{itemize}
\item Object detection and recognition module based on deep neural networks. They comprise several object detection modules (Mask R-CNN~\cite{he2018mask} and UOIS~\cite{UOIS}) and prompt-based classification using pre-trained CLIP~\cite{clip}. We also designed special prompts to enhance the performance of pre-trained CLIP for WRC2020 tidy-up task. 

\item An image segmentation module was trained in a sim-to-real manner. We randomized several aspects of the environment in the simulator, such as robot configurations, drawer knob positions, and object sizes. 

\item Human pose recognition module to fulfill the human-in-the-loop task (``pass an object to the person shaking their hands''). 

\item Predicting grasp pose combining classic principal components analysis (PCA) and reinforcement learning in the simulators.

\item Grasp detection using a vision-based tactile sensor. 

\end{itemize}

It is worth mentioning that various data-driven methods have been proposed in various robot learning tasks, such as \red{locomotion~\cite{lee2020learning,yang2020data}, object grasping~\cite{joshi2020robotic}, and pick-and-place~\cite{berscheid2020self}}.
While these components are also crucial in the development of the tidying up robot, the development of each components alone is insufficient to develop a complete system that can operate well in real worlds. 
For example, the robot system should be able to handle various inputs (such as robot internal states, odometry, and camera inputs) that usually have drastically different frequencies, and be optimized with respect to the computational workload to speed up the entire system. 
\red{The notable difference between WRC and many existing robot learning tasks is that WRC needs to build the entire system rather than solving a specific task. }

\red{
In the context of general-purpose household robots, tasks involved in household robot systems are standardized as benchmarks and some robot competitions are held.
For example, ERL Consumer Service Robots Challenge~\cite{basiri2019benchmarking} standardize functionalities and tasks on environment recognition, human-robot interaction, and object manipulation in house environments.
Recently, RoboCup@Home Domestic Standard Platform League (DSPL) adopts a similar rule to the WRC2020 tidy-up task~\cite{yamamoto2019human,matamoros2018robocup,matamoros2019trends}.
However, many studies based on these competitions on household service robot systems are rather biased towards the system integration for certain benchmarks than discussing the generalization and have explored separately from the robot learning studies,}
\blue{ while some recent solutions of these competitions partly leverage learning modules (e.g. off-the-shelf object recognition modules)~\cite{lima2019socrob, van2019tech, song2021robocup}.}
\blue{In the WRC2020 competition, we showcased a service robot system which extensively utilized data-driven modules for generalization and adaptation in household environments durable for real-world deployment.}

To utilize data-driven approach in WRC2020 task, it is important to consider how to design the entire system and how to separate the complex system into trainable modules as well as how to train each module. 
However, many data-driven approaches are only tested on benchmarks that does not require entire system design, or detailed design of the entire system are not comprehensively discussed\blue{~\cite{sac-rss,mahler2019learning,shridhar2022cliport}}. 
This paper describes an entire system for tidying up robot that works in real space using data-driven approach\blue{~(e.g. data collection, model selection, learning, and communication among modules)}, which might help practitioners to develop other data-driven robots.

The remainder of this paper is organized as follows.
Section~\ref{sec:overview} describes the principle in system design of the developed robot system and the overview of its software and hardware.
In Section~\ref{sec:recognition}, we present the modules used in the system for object and environment recognition, aiming to ensure robustness for deviation in the environment easy adaptation.
Section~\ref{sec:planning_manipulation} describes motion planning for navigation and manipulation reflecting the result of forementined recognition modules.
Subsequently in Section~\ref{sec:experiment}, we present the result of integrated experiment that took place in WRC2020 competition and discuss the current limitation and the plan to extend our robot system.

\section{System Overview}
\label{sec:overview}

This section describes the entire system for tidying up the room task in the WRC2020. 
Section~\ref{sec:task} describes the tidy up task in this competition in detail.
A major difficulty for service robots to work in the household environment is the significant variety in the real environments; there are numerous edge cases to consider because of the unknown environment and the high degree of freedom in behavior. 
In this case, we solely rely on data-driven approach, particularly deep learning, to handle the issue. 
Section~\ref{sec:DNN} describes how we useed the data-driven approach, especially deep learning, in our systems. 
Furthermore, we discuss the importance of the adaptation strategy in the data-driven approach. 
While the deep learning has exhibited  great performance in various applications, its utilization is still limited in robot systems. 
Difficulty in the application of deep learning in complex robot system attributes the huge computation that the deep learning based modules typically require, and without proper system design, the system throughput is significantly impaired. 
In Section~\ref{sec:statefull}, we describe object-centric state-full system (implemented with ``object manager'') where the system stores and updates the result of perception about the world in the object-level.
The object manager asynchronously distributes the most up-to-date information to each module, and prevents the deep learning modules from bottlenecking the computation.  
Finally, we describe the software and hardware architecture at the end of this section.

\subsection{Task Description of WRC2020 Partner Robot Challenge}
\label{sec:task}
\begin{figure}[t]
    \centering
    \includegraphics[width=0.75\linewidth]{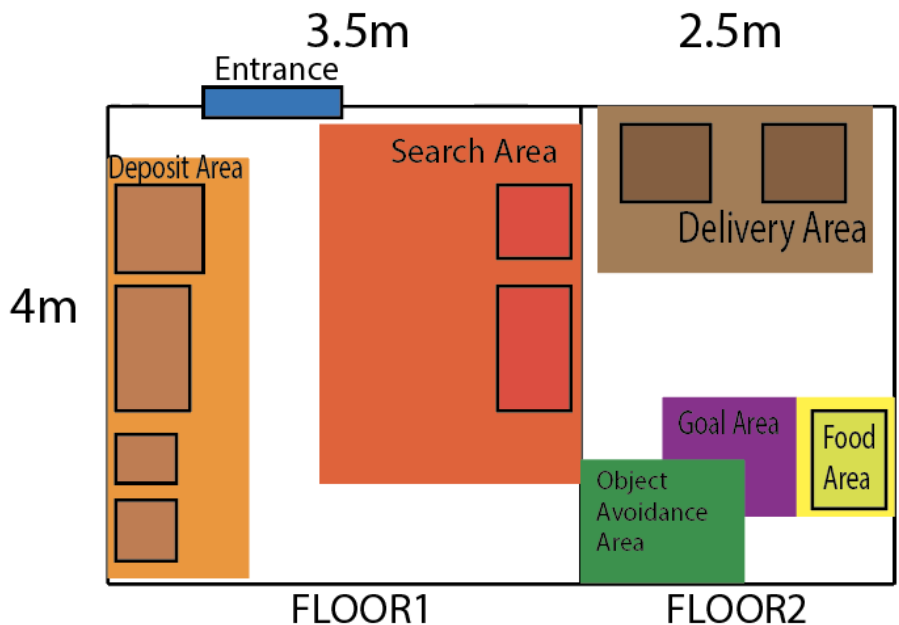}
    \caption{Floor map. In task 1, there are approximately 20 items on the search area in Floor1.
    The robot attempts to pick up these objects and place them into designated deposit area.
    After finishing task 1, the HSR moves to Floor2 through object avoidance area to goal area.
    Subsequently, the robot picks up the specified food object from the shelf.
    Lastly, the robot is asked to deliver the object a person waving a hand in the delivery area.
}
    \label{fig:map}
\end{figure}

In task1 of the competition (15 minutes), the robot picks up the object scattered on the floor and tables and place them in the target area specified by their object category in Floor1.
In the task2 (5 minutes), the robot moves to the neighboring room (Floor2) while avoiding small objects scattered in the floor, where it is asked to pick up the specified objects from the shelf and deliver it to a person waving the hand to the robot in delivery area.
Figure~\ref{fig:map} illustrates the layout used in WRC2020.

\begin{figure}[t]
    \centering
    \includegraphics[width=0.9\linewidth]{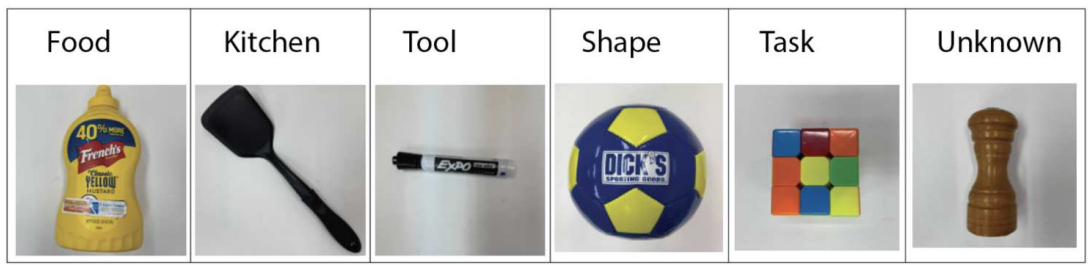}
    \caption{Objects' categories~(examples).~\cite{calli2015}  All objects are categorized under ``Food'', ``Kitchen'', ``Shape'', ``Tool'', ``Task'', and ``Unknown'', except for a boss item. "Unknown" is also categorized to other categories. "Unknown" object is not revealed just before the trials. }
    \label{fig:ycbs}
\end{figure}

In the task 1, the robots are required to quickly store objects scattered on the floor and the desk into the designated places.
Regarding tidying up in Floor 1, each item is categorized as shown in Figure~\ref{fig:ycbs}, and each category has a designated place to be stored.
The scores are added regardless of where the items are placed, but the score is higher if the items are stored in the designated place.
If the robot bumps into an object on the floor while moving, the score of the object it tries to store away afterwards will be minus.
Some items, such as tableware and pens, have a designated storage direction, and a ``boss item'' has higher points than other objects (in the WRC2020 tournament, the boss item was a golden spear).
Additionally, if the robot opens the three shelves in the deposit area, the bonus points are added to the total score.

In task 2, the robots are expected to be able to avoid obstacles in tight spaces, plan grasps in tight spaces, and interact with humans.
In concrete, after finishing task1, the robot has to avoid hitting the small items and navigate to the goal area in front of the shelf in Floor2~(task2a).
Subsequently, the robot delivers the target objects on the shelf in food area to a person in the delivery area.
There are two judges in Floor2, one is waving the hand, and the robot is asked to deliver the item to the waving person~(task2b).
WRC2020 rule book~\cite{wrsrule} explains the rules and scoring system in detail.

\subsection{Deep Neural Networks based System}
\label{sec:DNN}
One of the difficulties in introducing service robots into the home is that there are many possible edge cases because of the unknown environment and the high degree of freedom in behavior.
Traditionally, the developer needs to create program one-by-one to handle each edge case that they encounter. 
Recently, the advances of deep learning in various application areas (such as computer vision and natural language processing) has led to the introduction of various deep learning models for various robotics tasks.
Unlike in a manual programming approach, a deep learning based approach learns from the data, and can adapt to new edge cases by adding appropriate data.

\begin{figure}[t]
    \centering
    \includegraphics[width=0.9\linewidth]{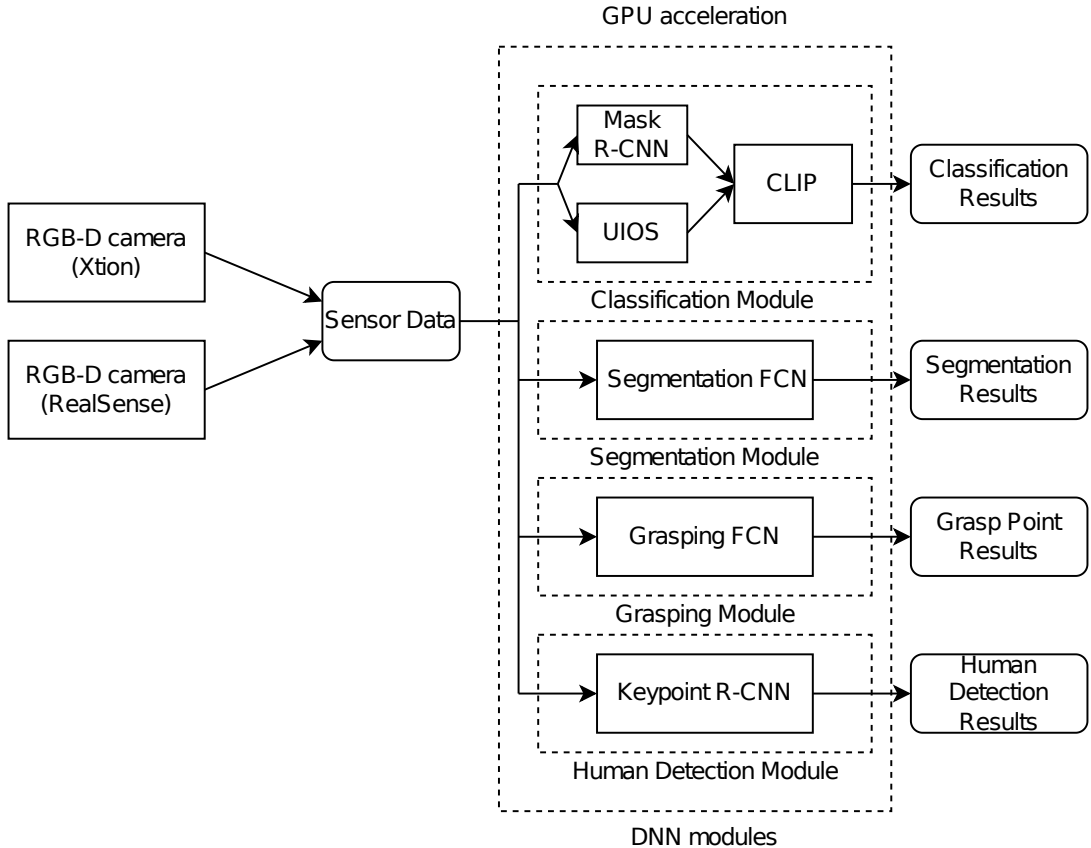}
    \caption{The proposed DNN-based module configuration. The module receives RGB-D image from Xtion and RealSense mounted on HSR.
    The corresponding sensor data is transmitted to each DNN-based module, as shown in Table~\ref{tab:dnn_table}.
    The process of each module operates at a sufficient frequency owing to parallelization and GPU acceleration.
}
    \label{fig:dnn_node}
\end{figure}

The proposed robot system makes use of various deep learning based recognition modules, and by randomizing the environment during training, we successfully developed a robust system that can respond to changes in the environment without any hard coding.
Table~\ref{tab:dnn_table} lists the correspondence between the model and the input data used in the system.
However, such powerful recognition models require a long time for inference and can be a bottleneck for real-time execution.
To solve this problem, we partitioned each deep learning module (e.g. object detection, classification, grasp proposal, motion planning, etc.) into ROS nodes as shown in Figure~\ref{fig:dnn_node}.
Each of these nodes operates in parallel and communicates asynchronously, which guarantees the speed and performance of the system.

\begin{table}[htbp]
    \centering
    \tbl{Type and purpose of the model used and corresponding inputs.}{
    \begin{tabular}{c|c|c}
        Model Name & Input Data & Purpose\\
        \hline\hline
        CLIP~\cite{clip} &  RGB image from RealSense and Xtion & Classificaion\\
        Keypoint R-CNN~\cite{he2018mask} &  RGB image from Xtion & Human Detection\\
        Mask R-CNN~\cite{he2018mask} & RGB image from RealSense and Xtion & Object Detection\\
        UOIS~\cite{UOIS} & RGB-D image from RealSense and Xtion & Object Detection\\
        Segmentation Model~(Section~\ref{sec:sim2real_recognition})  & Depth image from Xtion & Segmentation\\
        Grasp Model~(Section~\ref{sec:grasp_pose_prediction}) & RGB-D image from RealSense and Xtion & Grasp Point Calculation
    \end{tabular}}
    \label{tab:dnn_table}
\end{table}

One of the challenges in operating a human support robot in a home is to adapt itself to different environments in each home.
However, it is difficult to develop a robot system that learns how to operate in all possible environments in advance.
The proposed robot system has a particular advantage in terms of adaptability to the environment for future operation in real life.
More specifically, in the proposed system, data is collected each time the robot is operated, and the data collected is used to fine-tune each module, which creates a model that can adapt to each house environment during deployment.
For example in the competition of WRC2020, 
because the lighting conditions are much different from those of our laboratory, the performance of recognition module was slightly impaired.
However, by simultaneously collecting data during the four rounds of round-robin competition and fine-tuning the classifier, we succeeded in surpassing the performance of any round-robin competition in the semifinals.

\subsection{Object-centric State-full System} \label{sec:statefull}

Creating stateless modules is a straightforward implementation to utilize DNN-based modules in robot system. 
However, this design choice may significantly decrease the throughput of the entire system in the presence of DNN-based modules (such as object detection, key-point detection, and classification), because they require far more computational time than other modules, such as motion planning even with appropriate GPU accelerations and become the main bottleneck of the system.
This is a problematic especially when some module depends on the output of the DNN-based module.

To avoid the reduction of entire throughput, we propose to use a stateful, object-centric design. 
The design is based on the idea of using parallelization and distributed processing to enhance computational efficiency.
In summary, a module in the proposed system (called ``Object manager'') caches the outputs of the recognition module in the object-level, and each module communicates through the object manager. 
To be more specific, the object manager records information about each object (e.g., the detected location of the object, classification results) obtained from the deep learning models.
By recording the information of each object, such as location and classification results in the object manager in advance, and utilizing the stored results, we can reduce computation and achieve efficient operations.
For instance, the robot can move directly to the next target object after placing the object, without having to perform object recognition again.

\begin{figure}[t]
    \centering
    \includegraphics[width=0.9\linewidth]{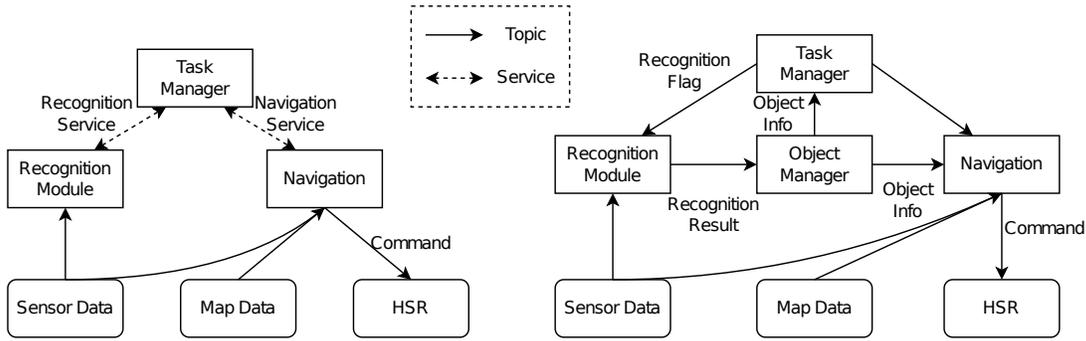}
    \caption{Brief overview of the proposed robot system. The left side of the figure is our initial stateless system design, and the right side of the figure is the final system design, the stateful and object-centric one. Owing to asynchronous processing and the object manager with ability to store recognition results, the right system could reduce throughput degradation even when using computationally demanding DNN modules.}
    \label{fig:system_compare}
\end{figure}

Figure~\ref{fig:system_compare} presents a brief overview of the designed robot system.
The left side of the figure shows the initial stateless system design, and the right side of the figure shows the final system design, which is stateful and object-centric.
In the stateless system design on the left, 
the task manager calls recognition module requiring huge synchronous computations, resulting in a significant drop in throuput.
Meanwhile, in the stateful design on the right, the recognition module operates asynchronously with the task manager, and the recognition results are stored in the object manager accordingly, so the task manager does not have to call the recognition module when using the recognition results.
Furthermore, in the right system, if necessary, the task manager can stop the recognition module through the recognition flag. This enables efficient processing even when the GPU resources are low.

\begin{figure}[t]
    \centering
    \includegraphics[width=0.75\linewidth]{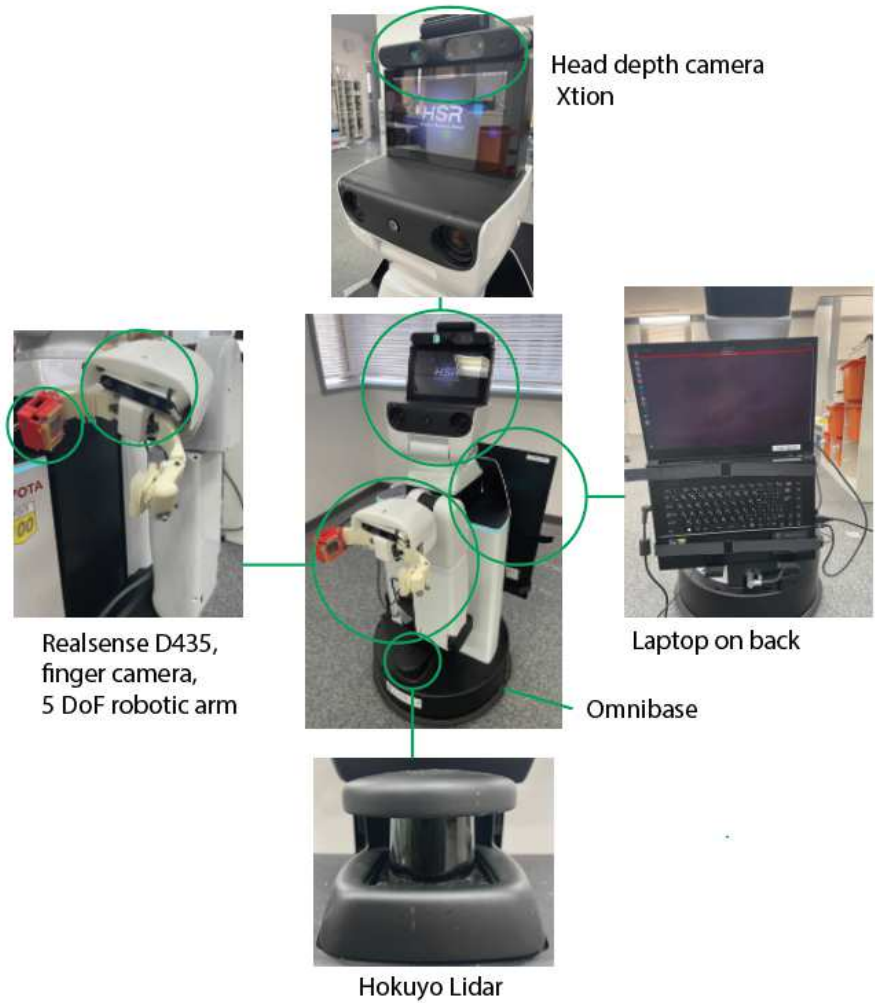}
    \caption{Customized HSR (Human Support Robot). HSR is a mobile manipulator designed to support human activities in daily life, such as nursing care and housework. It can pick objects under 500g and designed for safety in human interaction. In the competition, we focused on clean up tasks with HSR. The team replaced a hand RGB camera for the Realsense D435 and one of the fingertips for the tactile sensor module. }
    \label{fig:hsr}
\end{figure}

\subsection{Software} \label{sec:software}
We adopted the method proposed by L. El Hafi et al.~\cite{lofti, lofti2022} that configures a software development environment in a Docker container.
Using Docker has several advantages, such as easy management of module versions and the ability to rapidly develop the same environment on different machines.
In addition, unlike VirtualBox and other virtual machines,  Docker operates on the kernel of the host machine, so it is as fast as running on it.
It is crucial when executing systems that require computation, such as deep learning inference.
The proposed software development environment is based on Robot Operating System (ROS)~\cite{ROS} on Ubuntu.
Owing to the asynchronous communication feature provided by ROS, multiple modules can be processed efficiently in parallel. ROS also makes it easier to develop distributed systems, thus enhancing the stability of the system.

\subsection{Hardware} \label{sec:hardware}
Toyota's HSR (Human Support Robot)~\cite{yamamoto2019}, a mobile manipulator, was used for the robot, which has a motor, encoder, a 6-axis force sensor in the wrist, a Lidar in the foot, an Xtion depth camera~\footnote{http://xtionprolive.com/asus-xtion-pro-live} in the head, a stereo camera, and a microphone as external sensors.
As an additional sensor, a monocular wide-range RGB camera at the tip of the hand was replaced with a Realsense D435~\footnote{https://www.intelrealsense.com/depth-camera-d435/} depth camera (Figure~\ref{fig:hsr}). In addition to the internal processor, we used ROS~\cite{ROS} to communicate with a back-mounted laptop. For the laptop, we used an msi GS66 STEALTH with RTX3080 mobile 16GB \footnote{https://us.msi.com/Laptop/GS66-Stealth-10UX/Overview}.

\begin{figure}[t]
    \centering
    \includegraphics[width=0.9\linewidth]{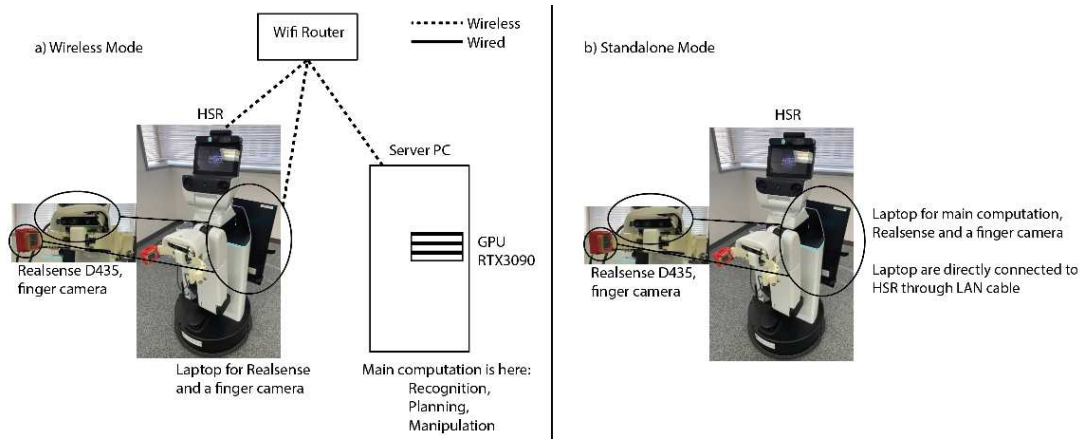}
    \caption{Wireless and Standalone mode. The wireless mode can use external powerful computational resources via a wireless network. The standalone mode computes all processes in the laptop installed on the back of HSR. This mode is much useful when the WiFi connection is weak. Our team picked standalone mode in the competition because WiFi bandwidth was limited to about 30 Mbps.}
    \label{fig:mode}
\end{figure}

For this competition, we created two modes of operation:~\red{wireless mode and standard mode}.
In the~\red{wireless} mode, the HSR is configured to operate in a high-speed WiFi environment ($\sim 1$ Gbps), and the GPU-intensive calculations are executed on an external PC and the results are transmitted to the HSR.
In addition, the most important communications were the depth camera images of \red{the hand and head}.
\red{The~\red{standalone} mode is useful when WiFi connection is weak, because all of the computation is performed on the backpack PC only and does not require wireless connection.
In the~\red{standalone} mode, multiple machine learning models were computed on a 16GB GPU~(Figure~\ref{fig:mode}).}

\red{
The mode selection is a trade-off between network speed and computation speed.
In the competition, the participants were allowed to use WiFi connection prepared by the organizer only, whose bandwidth was 30~Mbps at the maximum.
We chose the standalone mode, because we found that the bandwidth was too narrow for transmitting massive data (e.g. point cloud from two RGB-D cameras and LiDAR sensor), which caused severe delay in sensor data and critical errors in localization and navigation.
}

\section{Recognition}
\label{sec:recognition}

\red{
We aggressively utilized multiple DNN models aiming at generalization in recognition of objects~(Section~\ref{sec:object_recognition}), environment~(Section~\ref{sec:sim2real_recognition}), and human~(Appendix~\ref{sec:human_recognition}) in the competition arena.
}

\subsection{Object Recognition}
\label{sec:object_recognition}
In the WRC2020 task, the technology for fast and accurate recognition of objects is the fundamental technique for the overall task operation.
For example, if an object cannot be detected, or if detection is a slow process, it might be dangerous for collision with the object during navigation.
Moreover, if the object's exact shape cannot be obtained, the robot fails to calculate the correct grasping pose, resulting in grasping failure.
Furthermore, if the category of the object cannot be classified correctly, the robot cannot place the object in a defined location.
Thus, as object recognition is applied to various tasks such as grasping and navigation, it is crucial to estimate the pose, shape, and category of the object fast and accurately.

\begin{figure}[]
  \centering
  \includegraphics[width=0.75\linewidth]{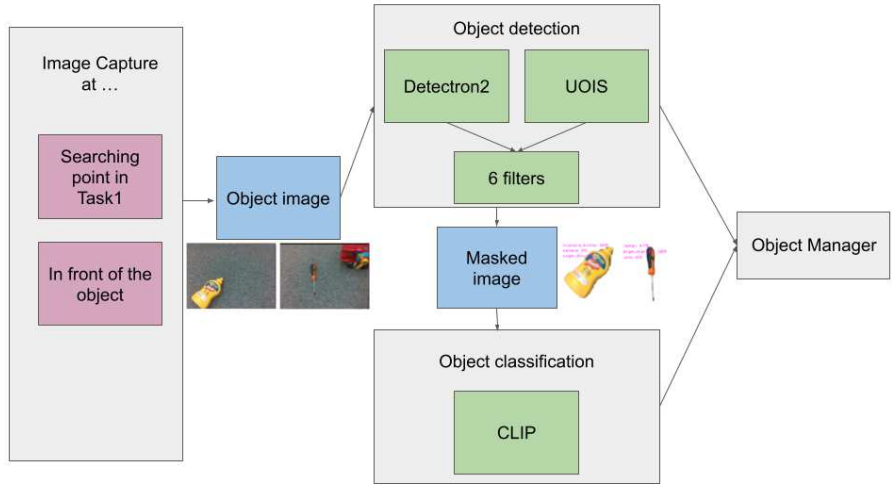}
  \caption{Object recognition flow. \red{Color images are captured with head-mounted or hand-mounted RGB-D camera and the masked images for each object in the captured images are created using object detection module. Subsequently, the masked images are classified using the CLIP classifier.
  In the case of the figure, the images of a mustard bottle and a screw driver on the floor are captured with the cameras. 
  The object detection module segments object area (namely, the area of the mustard bottle and screw driver) on the images and masks out other regions of images than those of the objects (e.g. the floors, toy airplane in the right image).
  }}
  \label{fig: recognition_flow}
\end{figure}

Object recognition system works in two situations, such as searching the objects and confirming the object to pick and place in tidying up in Room1~(task1).
To navigate the robot, it has to acquire the object's information about the pose and the category of the objects.
In the proposed system, before the robots starts picking the objects, it looks around the searching area using the head camera and recognizes the objects for searching.
The robot stands outside of the search area near the drawer, and divides the search area into three parts~(floor, lower table, and upper table), and images of each place is captured by the head camera.
After deciding the object to pick, the robot moves in front of the object and investigates it using hand camera from an angle above it for confirmation.
At this instant, the robot recognizes the object and plans how to grasp the object.
In picking the target object from the food area's shelf in task 2b, the robot photographs each row of the shelf. 
The hand camera moves in front of the center of each row of the shelf, and captures an image of the left side and the right side of each row.

Recognition system is divided into the detection part and classification part.
Owing to the two stages of object recognition module, during picking objects, the robot does not need to wait for the output from the object classification model, which usually involves extensive computations.

\subsubsection{Object Detection}
\label{subsubsec:object_detection}
\begin{figure}[]
  \centering
  \includegraphics[width=0.75\linewidth]{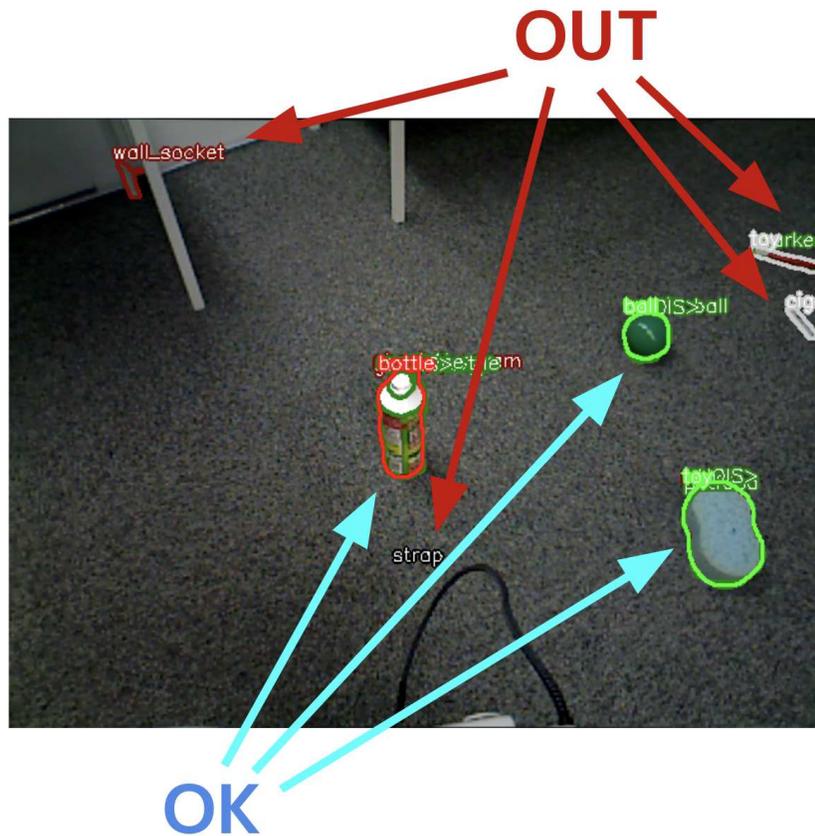}
  \caption{\red{Example of detected objects, each visualized with a colored outline based on results by hand-crafted filters. Green objects are pickable candidates. The far red object is out-of-bounds, the red bottle overlaps with a near-duplicate detection, and the gray objects are too close to the edge of the camera view.}}
  \label{fig:detection}
\end{figure}

In the object detection module, we used two object detection models, Mask R-CNN~\cite{he2018mask} in Detectron2~\cite{detectron2} and UOIS~\cite{UOIS}, as a combination.
Mask R-CNN in Detectron2 is a widely adopted object detection model and UOIS is robust for detecting unseen objects.
In the WRC2020 tidy-up task, because a lot of objects are placed in the nearest area, the system has to detect overlapping objects.
Even if assembling these two models, some furniture or floors are erroneously detected as objects.
The detected results were filtered to avoid misidentification~\red{using hand-designed rules}: excluding objects with too small or too big pixel areas, excluding objects whose labels are not in the valid list, excluding overlapped objects, excluding objects that are not in the task1 area, and excluding objects at the edge of the image, as shown in Figure~\ref{fig:detection}.
Although we can obtain the object label from these two detection models, the accuracy of classification is insufficient, which motivates combining another classification module as in Section~\ref{sec:object_clasification}. 
Figure~\ref{fig: recognition_flow} illustrates the object recognition workflow. Mask R-CNN and UOIS are primarily used for segmentation images.

\subsubsection{Object Classification}
\label{sec:object_clasification}
After applying the segmentation mask from detection module mentioned above to the original RGB images, the image is classified by CLIP~\cite{clip}, which is a recently proposed multi-modal deep learning model.
CLIP comprises three parts, text encoder, image encoder, and fully connected layers.
CLIP inputs the prompt, the text description of the object, and the picture, then outputs the probability of identification between the name and the picture.
The advantage of CLIP over other classification models is that it is easy to tune the model to the specific situation by only editing the prompt.
\red{This technique to improve performance using text prompts for clarifying the task has been proposed in a line of studies on large language models~(LLMs) and is called ``prompt tuning''\cite{brown2020language,lester2021power,zhang2021amortized,kojima2022large}.}
For task1 and task2, we prepared prompts for each object that does not directly use the label name, but the handmade descriptions, such as “red strawberry with green stem, a type of fruit” \red{as shown in Table~\ref{tab:prompts}}.
To adapt to the task situation, \red{we used pretrained CLIP model (ViT-B/32)~\footnote{Obtained from \url{https://github.com/openai/CLIP}.} and the only the fully connected layers after pretrained CLIP module were fine-tuned with cropped YCB image datasets~\cite{calli2015}}.
\red{
The fully connected layers are trained separately for task1 (the model is trained with objects of all categories) and task2b (the model is trained with only ``food'' objects), while the text prompts are shared between task1 and task2b.
}
Besides, considering the change in the light condition the floor pattern, we captured some images of objects which were placed on the same as competition during preparation day.
Those images were used to fine-tune the fully connected layers, and the process is very computationally light.
Data augmentation is also performed in each training to make the model robust to input noise.

\begin{table}
    \centering
    \tbl{
        \red{Known objects appeared in the competition and the corresponding prompts which we tuned. 
        For the task~2b, we trained object classifier with food objects only (denoted by $\surd$ in the table).}
    }{
    \begin{tabular}{c|c|c}
    Object Name & Prompt & \red{Used in Task~2b}\\
    \hline\hline
    chips can & a photo of a pringles can &  \red{$\surd$} \\
    master chef can & a photo of a blue coffee can & \red{$\surd$}   \\
    cracker box & a photo of a box of cheez-it &   \red{$\surd$} \\
    sugar box & a photo of a yellow box of Domino sugar &   \red{$\surd$} \\
    tomato soup can & a photo of a can of campbells tomato soup &   \red{$\surd$} \\
    mustard bottle & a photo of a yellow mustard &   \red{$\surd$} \\
    tuna fish can & a photo of a StarKist can tuna &   \red{$\surd$} \\
    pudding box & a photo of a box of brown chocolate pudding jello &  \red{$\surd$} \\
    gelatin box & a photo of a box of red jello &   \red{$\surd$} \\
    potted meat can & a photo of a can of spam &   \red{$\surd$} \\
    banana & a photo of a banana, a type of fruit &   \red{$\surd$} \\
    strawberry & a photo of a red strawberry with green stem, a type of fruit &  \red{$\surd$}  \\
    apple & a photo of a red apple, a type of fruit &   \red{$\surd$} \\
    lemon & a photo of a lemon, a type of fruit &  \red{$\surd$} \\
    peach & a photo of a yellowish peach, a type of fruit &  \red{$\surd$} \\
    pear & a photo of a green pear, a type of fruit &  \red{$\surd$} \\
    orange & a photo of a orange, a type of fruit &  \red{$\surd$} \\
    plum & a photo of a purple plum, a type of fruit &  \red{$\surd$} \\
    pitcher base & a photo of a blue pitcher & \\
    pitcher lid & a photo of a pitcher lid & \\
    bleach cleanser & a photo of a soft scrub bleach bottle & \\
    windex bottle & a photo of a windex spray bottle & \\
    wine glass & a photo of a wine glass & \\
    bowl & a photo of a red bowl & \\
    mug & a photo of a red mug & \\
    sponge & a photo of a sponge & \\
    plate & a photo of a plate & \\
    fork & a photo of a red fork & \\
    spoon & a photo of a red spoon & \\
    spatula & a photo of a spatula & \\
    key & a photo of a key & \\
    large marker & a photo of a expo marker & \\
    small marker & a photo of a expo marker & \\
    plastic bolt & a photo of a bolt & \\
    medium clamp & a photo of a clamp & \\
    card & a photo of a card & \\
    ball & a photo of a sports ball & \\
    rope & a photo of a rope & \\
    chain & a photo of a chain & \\
    foam brick & a photo of a brick & \\
    dice & a photo of a die & \\
    marbles & a photo of a marble & \\
    cups & a photo of a toy cup & \\
    peg & a photo of a wooden puzzle & \\
    toy airplane & a photo of a toy airplane & \\
    magazine & a photo of a time magazine & \\
    shirt & a photo of a black t-shirt & \\
    lego duplo & a photo of a lego & \\
    timer & a photo of a timer & \\
    rubiks cube & a photo of a rubiks cube & \\
    \end{tabular}}
    \label{tab:prompts}
\end{table}

\subsubsection{Evaluation}
\begin{figure}[]
  \centering
  \includegraphics[width=0.9\linewidth]{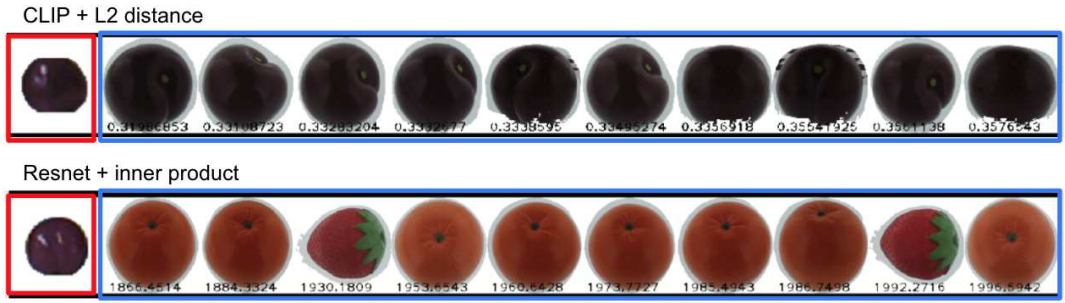}
  \caption{Classification results images for comparison between ResNet and CLIP.
  The upper image shows the CLIP result, and the lower image shows the ResNet result.
  The image surrounded by a red frame (a plum) is the recognized image and the images surrounded by a blue frame are images classified to be similar.
  }
  \label{fig:ResNet_clip}
\end{figure}

\begin{figure}[]
  \centering
  \includegraphics[width=0.6\linewidth]{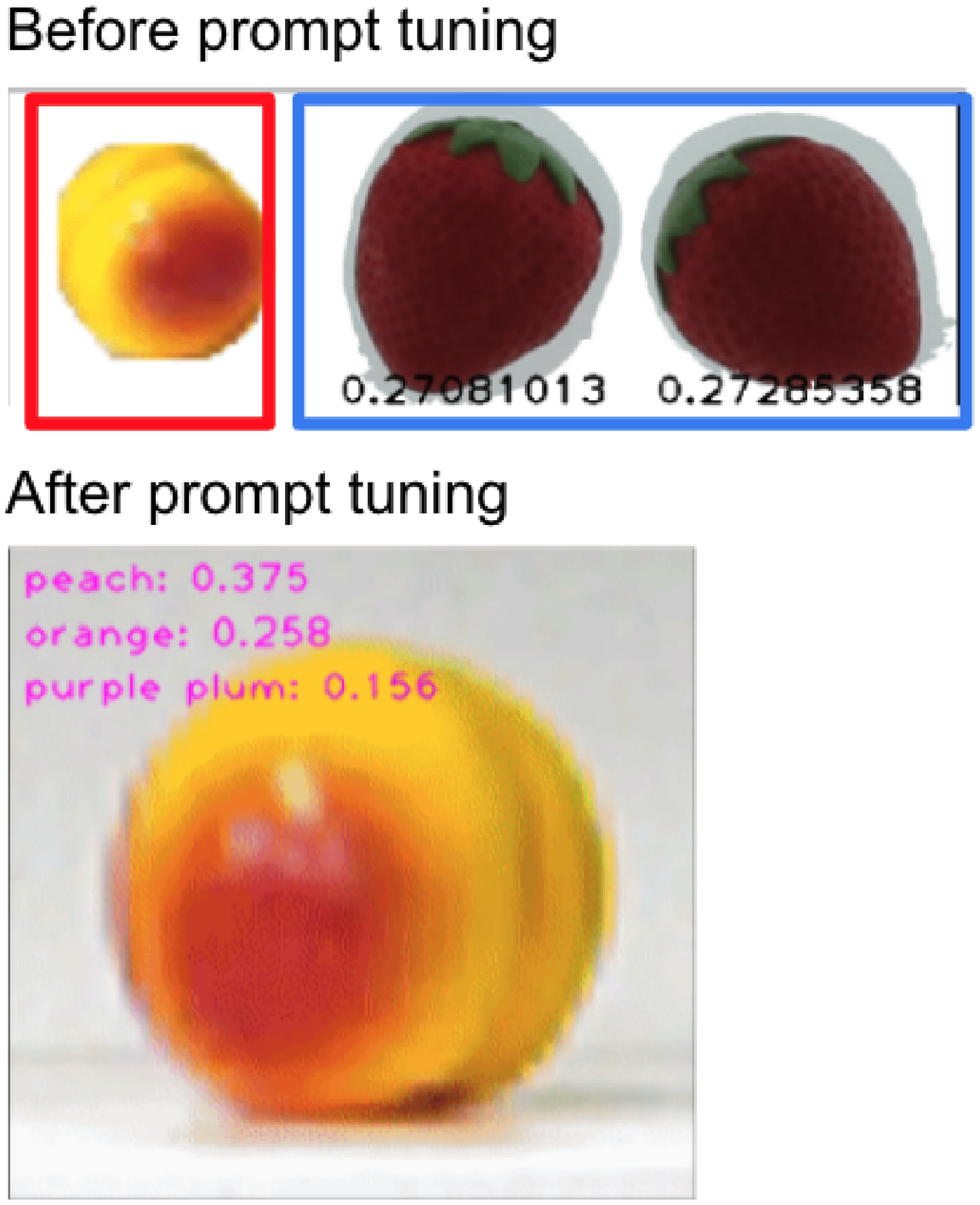}
  \caption{Comparison between with and without prompt tuning.
  The upper image illustrates a result of recognition without prompt tuning; the object to classify is a plum (surrounded with red frame), but the model erroneously outputs the nearest images (surrounded with blue frame) of strawberry, and recognizes the target object as a strawberry.
  On the other hand, the lower image illustrates the result with prompt tuning; the model correctly recognizes the image of a peach.}
  \label{fig:prompt_tuning}
\end{figure}

\begin{figure}
\begin{center}
    \subfigure[Successful case]{
    
        \includegraphics[width=0.5\linewidth]{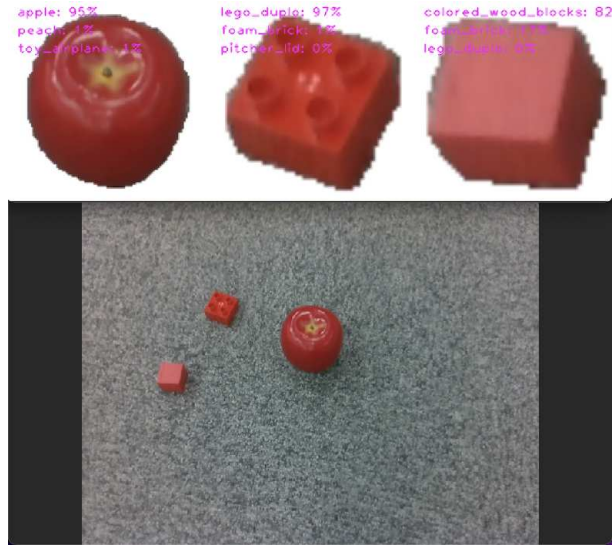}
    }
    \hspace{5pt}
    \subfigure[Failure case]{
    
        \includegraphics[width=0.5\linewidth]{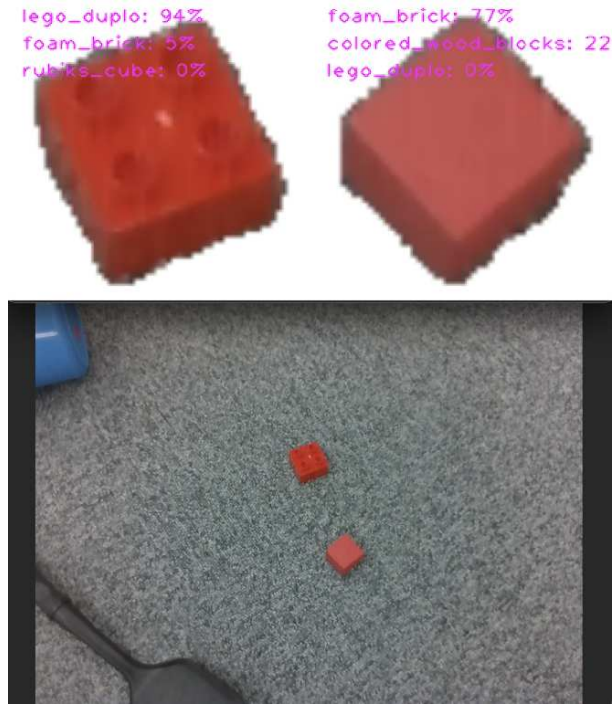}
    }
    \caption{\red{Examples of success and failure classification among similar objects: apple, Lego and wood block.In (a) our classifier can recognize similar color objects with subtle texture difference, but in (b), the colored wood block is recognized as a form brick (while Lego is classified correctly). }}
    \label{fig:apple-lego-wood}
\end{center}
\end{figure}

\red{
In addition to the CLIP model, we prepared the pretrained ResNet18 model\footnote{Obtained from~\url{https://github.com/pytorch/vision}}~\cite{resnet} as a baseline model, 
For the baseline ResNet model, the weight of ResNet module was fixed and we classified the object class of given images according to k-nearest neighbors of L2 distance of the output of ResNet modules.
For the CLIP model, we used inner product as a metric for object classification.
Compared to the baseline ResNet~\cite{resnet} model, CLIP exhibits more accurate classification ability as shown in Figure~\ref{fig:ResNet_clip}.
}
Before prompt tuning, CLIP classifier made mistakes the same way as the pretrained ResNet model; for instance, lemon was classified as strawberry.
However, the accuracy of classification increased after prompt tuning, as shown in Figure~\ref{fig:prompt_tuning}.
In detail, we made prompts including color information, such as ``blue coffee can''. 
Besides, the phrase ``the type of fruits'' improves the classification accuracy by recognizing strawberry as a strawberry, not an apple.
In addition to prompt tuning, background subtraction improved classification accuracy compared to only cropping.
In each training, the five types of data augmentations were applied to the training image, translating, scaling and rotating, dropping out the rectangular regions, changing hue, saturation and value, changing brightness, and contrast and blurring~\cite{albumentations}. 
\red{Figure~\ref{fig:apple-lego-wood}a illustrates successful classification among similar colored objects with the proposed recognition module. 
Despite these improvements, some objects were still identified incorrectly; for example ``Lego'' as ``nine peg hole test'', or ``colored wood block'' as ``foam brick''(Figure~\ref{fig:apple-lego-wood}b).
The performance gain can be attributed to the pretraining of CLIP, which is trained with massive and unrestricted paired (image and text) datasets collected from the internet. The dataset could contain some hints on the textural information on objects appearing in the competition.
}

\subsection{Recognition of World}
\label{sec:sim2real_recognition}
\subsubsection{Sim-To-Real Image Segmentation}

Since the layout of the room and positions of furniture are known beforehand, one may think of using this information as constants.
For example, one may carefully measure and calculate the (3D) position of the drawer knobs beforehand, and use those positions to grasp at that location in a completely open-loop fashion. However, this is a brittle and risky approach, primarily because numerous factors can cause a failure to pull the drawer, such as localization error, positions of knobs being slightly shifted, or drawers being too far in or out, causing missed grasps or unwanted collisions.

Instead of adopting this simple but risky approach, we implemented a perception module that can consistently estimate the state of the environment within view. 
A crucial component of this module is a single image segmentation model trained to recognize various objects and furniture in the room.
Specifically, a Fully Convolutional Network (FCN)~\cite{long2015fully} based on the DeepLab-v3 architecture~\cite{chen2017rethinking} was used to predict the class of each pixel of a given depth image.

The depth image, which was obtained from the head camera of the robot, had a resolution of $640\times480$ pixels. Units are in meters and missing values were assigned a value of zero.
We deliberately decided to avoid using color (RGB) for this model to circumvent the challenge of generalization over different lighting, textures, and other visual phenomena.

The following lists the classes that the model is trained to predict.

\begin{itemize}
  \item Background
  \item Wall
  \item Pickable object
  \item Shelf
  \item Left bin
  \item Right bin
  \item Drawer frame
  \item Bottom drawer
  \item Bottom drawer knob
  \item Left drawer
  \item Left drawer knob
  \item Top drawer
  \item Top drawer knob
  \item Miscellaneous drawer
  \item Tall table
  \item Long table A
  \item Long table B
  \item Left tray
  \item Right tray
  \item Left container
  \item Right container
\end{itemize}

\begin{figure}[]
  \centering
  \includegraphics[width=0.6\linewidth]{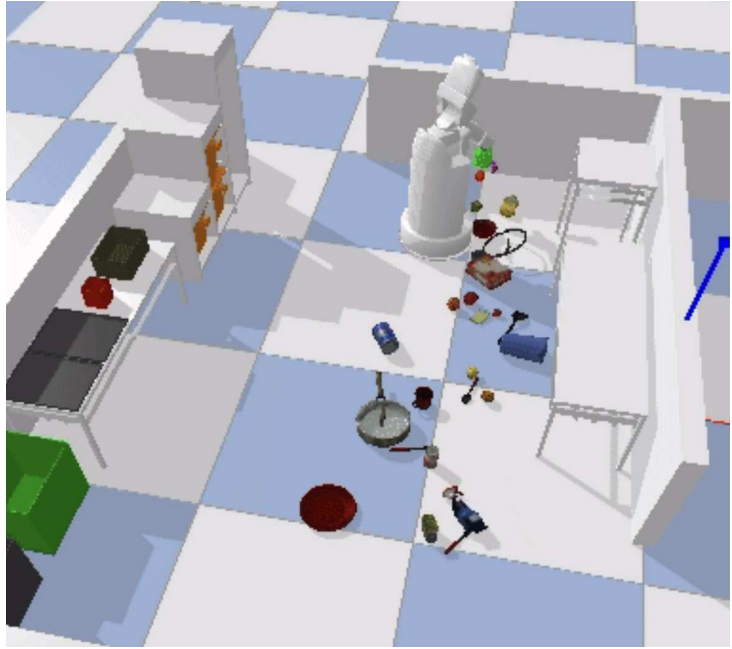}
  \caption{Simulation environment used for image data generation. The room layout and most of the furniture reflect the real competition environment.}
  \label{fig:sim}
\end{figure}

Obtaining a sufficient amount of human-annotated training data can become expensive in terms of effort, time, and monetary cost. Instead, we adopted a sim-to-real technique by generating synthetic image data and annotations from a simulation. The developed simulation uses PyBullet~\cite{coumans2021} for both physics and rendering but many of the assets, including the URDF of the HSR and furniture models, were ported from an existing Gazebo~\cite{Koenig2004} simulation. The resulting simulation is depicted in Figure~\ref{fig:sim}. To train a robust model, the synthetic depth images were rendered and applied with noise during training to mimic the noisiness of the real depth camera. Scenes were also generated with randomization. The following lists what was randomized for data generation.

\begin{itemize}
  \item Robot configuration
  \item Wall height and thickness
  \item Shifted positions and rotations of furniture
  \item Drawer knob position, rotation, and shape
  \item Drawer position, including open/close state
  \item Shape and size of trays and containers
  \item Presence of miscellaneous drawers
  \item Number, poses, shapes, and sizes of pickable objects
\end{itemize}

Variation in shapes were made by sampling meshes from ShapeNet~\cite{chang2015shapenet}.

\begin{figure}[]
  \centering
  \includegraphics[width=0.6\linewidth]{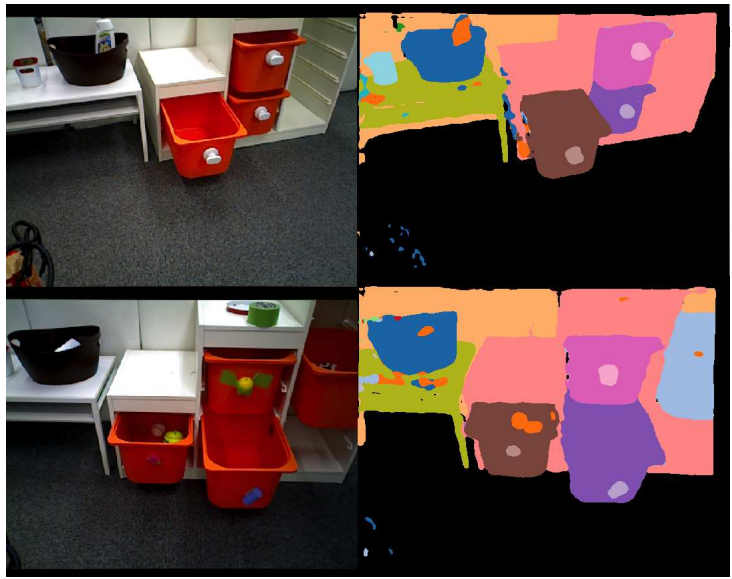}
  \caption{Two samples of model predictions on real data. The model can generalize to varying viewpoints, positions, and shapes.}
  \label{fig:sim_knobs}
\end{figure}

\begin{figure}[]
  \centering
  \includegraphics[width=0.6\linewidth]{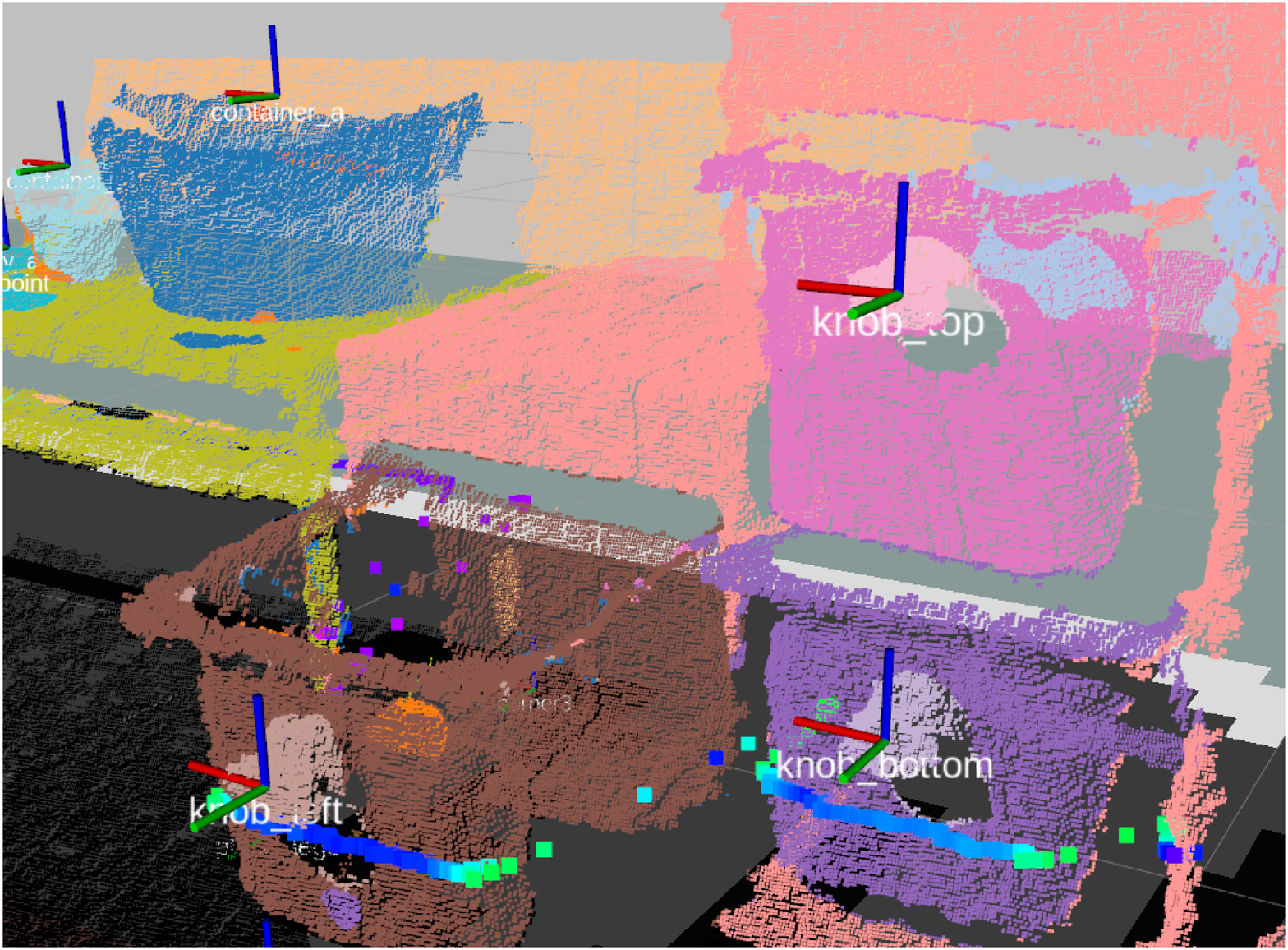}
  \caption{The perception system processes the model predictions with depth information to accurately estimate the state of the environment, such as the positions of drawer knobs.}
  \label{fig:knobs}
\end{figure}

\subsubsection{Evaluation}

Example predictions of the resulting model is presented in Figure~\ref{fig:sim_knobs}. The figure demonstrates its robustness, as it can recognize objects even when the camera viewpoint or shapes and of drawer knobs change. These predictions are used with depth information to obtain segmented point clouds which can then be used to estimate information, such as object positions, as shown for drawer knobs in Figure~\ref{fig:knobs}.

\section{Object Manipulation and Motion Planning}
\label{sec:planning_manipulation}
\subsection{Object Manipulation}
\subsubsection{Grasp Pose Prediction}
\label{sec:grasp_pose_prediction}

\begin{figure}
\begin{center}
    \subfigure[PCA]{
    \resizebox*{4cm}{!}{
        \includegraphics[width=0.3\linewidth]{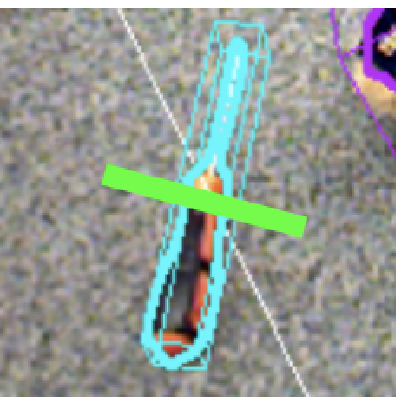}
    }}
    \hspace{5pt}
    \subfigure[\red{Multi-Object Multi-Grasp}~\cite{chu2018real}]{
    \resizebox*{4cm}{!}{
        \includegraphics[width=0.3\linewidth]{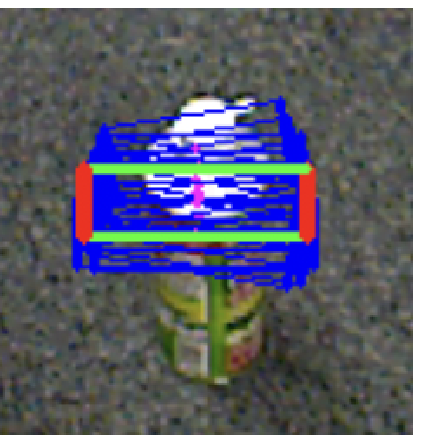}
    }}
    \hspace{5pt}
    \subfigure[GraspNet~\cite{mousavian20196}]{
    \resizebox*{4cm}{!}{
        \includegraphics[width=0.3\linewidth]{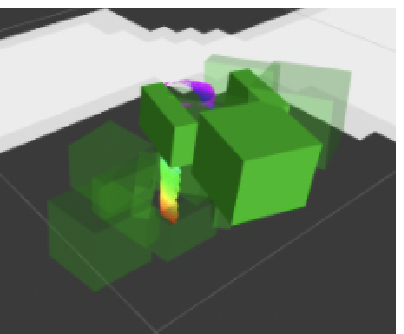}
    }}
    \hspace{5pt}
    \subfigure[GGCNN~\cite{morrison2020learning}]{
    \resizebox*{4cm}{!}{
        \includegraphics[width=0.3\linewidth]{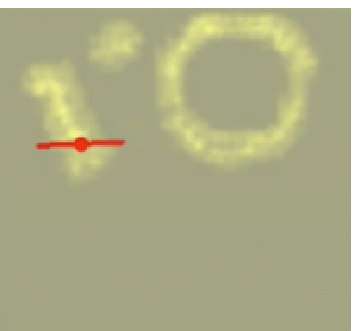}
    }}
    \hspace{5pt}
    \subfigure[Grasp FCN]{
    \resizebox*{4cm}{!}{
        \includegraphics[width=0.3\linewidth]{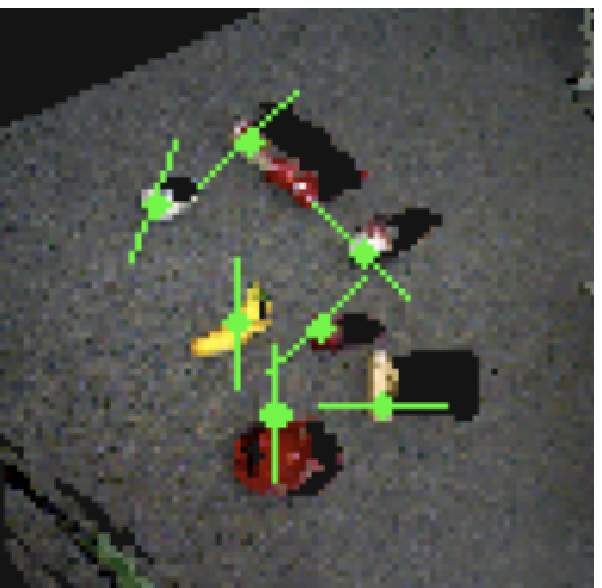}
    }}
    \caption{
    Visualizations of the attempted grasp estimation methods. Both PCA and GraspNet take as input a point cloud of a single object, but PCA calculates a 4 degrees of freedom (DoF) grasp while GraspNet can produce 6DoF grasps. Both the grasp detector and GGCNN take as input a depth image but the detector outputs bounding boxes while GGCNN outputs a grasp quality heatmap. The Grasp FCN also outputs a heatmap but takes as input a more view-invariant heightmap.
    \red{See Appendix~\ref{app:other_grasp} for the method illustrated in (b)-(d).
    }
    }
    \label{fig:grasp_methods}
\end{center}
\end{figure}

\begin{figure}[]
  \centering
  \includegraphics[width=0.75\linewidth]{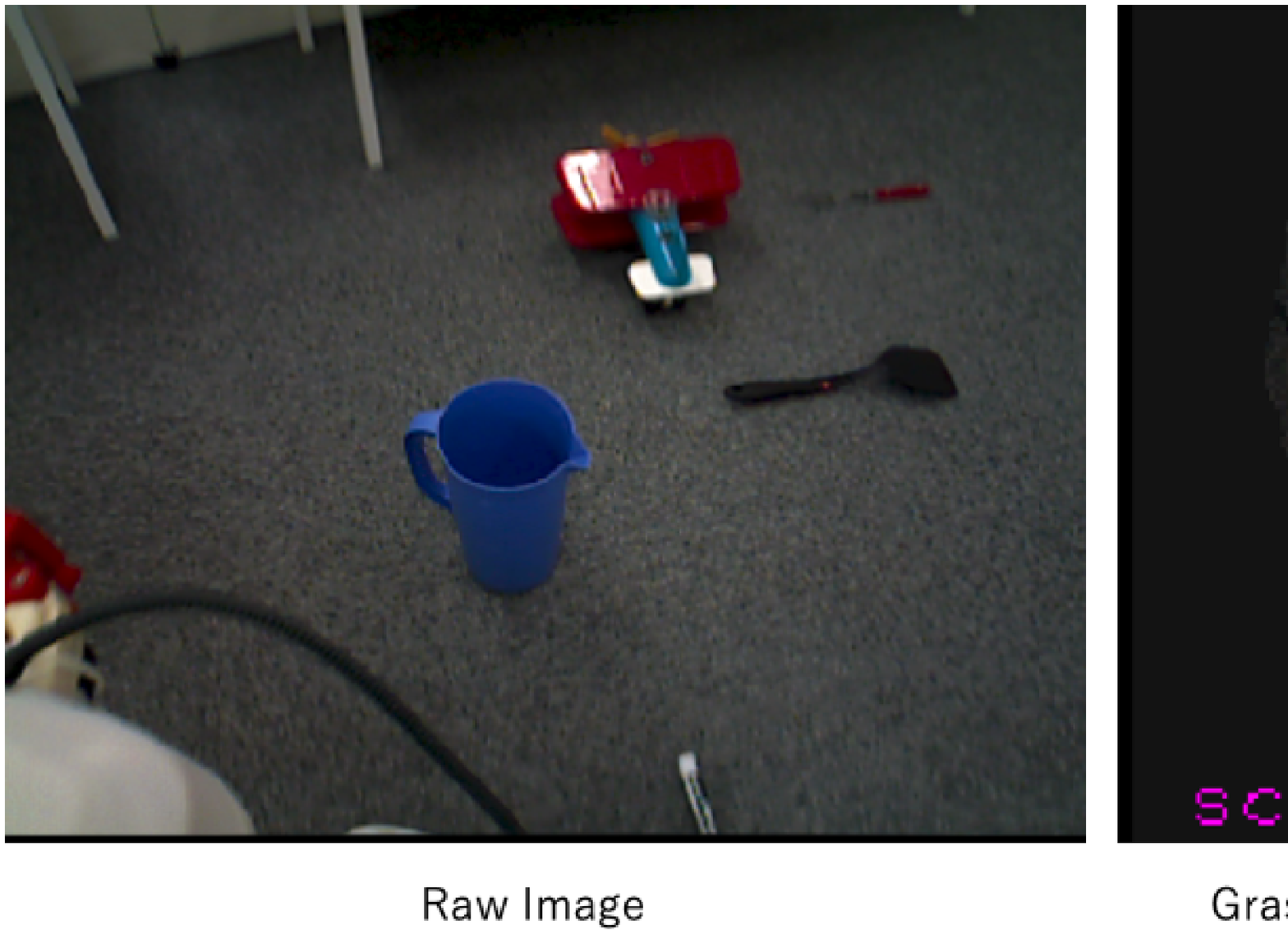}
  \caption{\red{Left: RGB image from a head camera. Grasp FCN generates grasp candidates with the quality score.
  Right: The red line illustrates the pose with highest score, and the green illustrates other candidates.}}
  \label{fig:fcn_grasp}
\end{figure}

Grasping becomes an important component of the robotic system to reliably place the objects scattered around the room at their assigned locations.
At the same time, grasping is challenging because different objects in different configurations can afford different stable grasp poses. 
\red{
We attempted various grasp planning methods as illustrated in Figure~\ref{fig:grasp_methods}.
In the main text, we describe the methods that we mainly used in the competition.
See Appendix~\ref{app:other_grasp} for the rest.
}

\paragraph*{\red{\textbf{PCA}}}
The first and most simple method we attempted was using principal components analysis (PCA). Given the point cloud of a target object, the points were projected to the XY plane and PCA with 2 components used to fit this data. A grasp was calculated so that the gripper is positioned above the center of the object and the fingers are aligned along the 2nd component of PCA. Intuitively, this results in grasps that are perpendicular to the longer axis, which is useful for estimating stable grasps of long objects, such as a banana. One assumption that this approach makes is that objects can be grasped in a top-down fashion, which is true in most cases in Task1.
In fact, we found this method to be the most reliable over the course of development.

\paragraph*{\red{\textbf{Sim-to-Real Transfer}}}
We attempted a custom DNN-based model which was trained in the simulator described in Section~\ref{sec:sim2real_recognition}. The model architecture and training method is largely based on~\cite{zeng2018learning}, a reinforcement learning method. The model takes a heightmap as input which can be generated by projecting a point cloud onto the XY plane, and outputs a map of Q-values, which corresponds to the likelihood of success for a grasp at the pixel location. Possibly owing to the resolution of the input heightmap, we found this method to lead to grasps that were occasionally off by a few centimeters.

\begin{figure}[]
  \centering
  \includegraphics[width=0.75\linewidth]{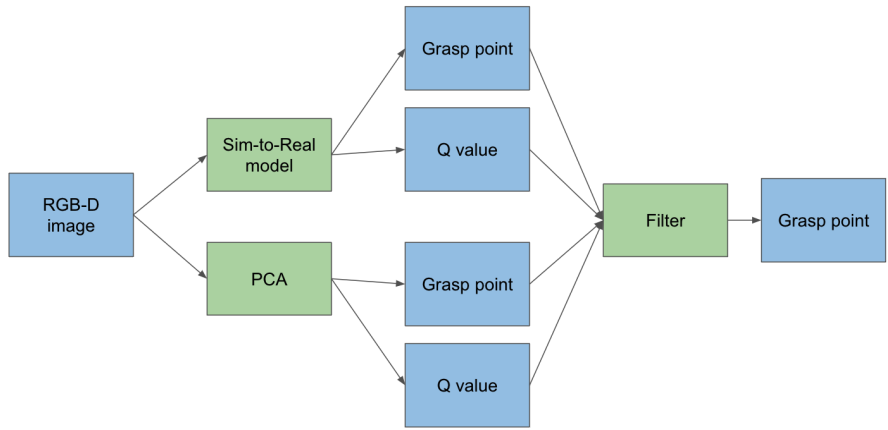}
  \caption{Grasp pose prediction flow}
  \label{fig:grasp_prediction}
\end{figure}

Ultimately, we opted to use the last method for generating grasp candidates~(Figure~\ref{fig:fcn_grasp}) and combined it with the PCA method.
The grasp pose was selected depending on the Q-value of the model~(Figure~\ref{fig:grasp_prediction}).
If the Q-value was above a set threshold, meaning that it was very confident, the model was used, else, PCA was used. Since we found PCA to be sufficient for most objects as they are simple and \red{convex}, this threshold was set very high.

\subsubsection{Grasp Detection with Hand-made Tactile Sensor}

The judgment of the success or failure of the grasp is crucial for the robot to perform the pick and place operation correctly. 
In HSR, it is relatively easy to implement the grasp detection according to the angle of the gripper and the 6-axis force sensor on the wrist.
However, it is difficult to detect a successful grasp of a light object, such as a pen or a key which appeared in the WRC2020 competition.

\red{We created a tactile sensor based on~\cite{Yamaguchi2016}, which is composed of black painted lead balls, transparent gels, acrylic board, a fish-eye camera, and 3D printed parts.} 
When force is applied to the sensor, the gel deforms and the lead balls move. The camera could capture the markers' movement and use it as the sensor value.
The advantages of the proposed tactile sensor is the ease of fabrication and the affordability of the amount of material.
The fabrication method of the sensor and the 3D printer parts are available as open-source, and the materials are relatively cheap~(about 60 USD) and easy to purchase.
For simplification of the fabrication process, we replaced a hand-made gel with a pair of non-slip sheets of furniture and place markers between the two sheets.

By using this sensor, we detected the grasping state of HSR.
There are three states, ``open'', ``objects in hand'', and ``nothing in hand''.
These three states are categorized by marker positions.

\begin{figure}[t]
    \centering
    \includegraphics[width=\linewidth]{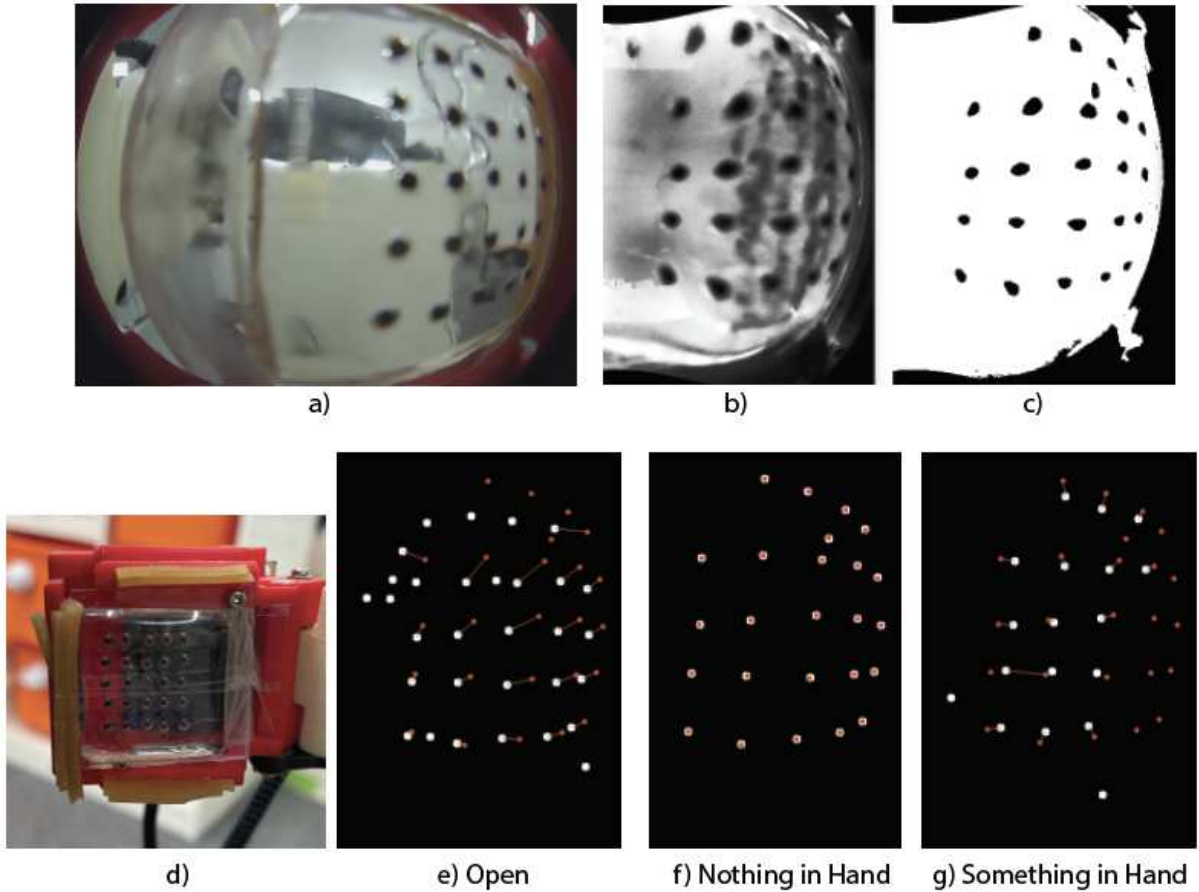}
    \caption{\red{Grasp detection mechanism.
    (a) Raw image taken by the finger camera.
    (b) Gray image of the image (a).
    (c) Binarized image by OTSU's method\protect\footnotemark.
    (d) Fingertip attached to HSR hand. The nails are made of rubber and lead balls are embedded into the gel.
    (e-g) Examples of marker positions.
    ``Open'' and ``Nothing in Hand'' images are registered, and when the markers positions deviates from both registered positions, it is detected as ``Something in Hand''.}}
    \label{fig:markers}
\end{figure}

For the calibration of marker positions and grasping state, we registered the initial markers positions in open and closed state.
The marker positions in the open and closed states were registered using the Blob detector\footnote{\url{https://docs.opencv.org/3.4/d0/d7a/classcv_1_1SimpleBlobDetector.html}} from a binalized camera image, as in Figure~\ref{fig:markers}c.

\red{Figure~\ref{fig:markers} illustrates how the marker positions have been registered and processed.
In Figure~\ref{fig:markers}e-g, the detected markers positions by Blob detector are illustrated with white dots.
For comparison, the marker positions of registered ``nothing in hand'' is illustrated by red points in image e-g.
In Figure~\ref{fig:markers}f, white dots and red points are almost in the same positions.
In the case of ``open'' state (Figure~\ref{fig:markers}e), the number and arrangement of white dots are different from the registered red dots, which means the Blob detector cannot detect the lead points correctly since the light comes into the finger camera from open gripper.
In the case of ``something in hand'' state (Figure~\ref{fig:markers}g), the white points are translated from red dots, which means that some object in the hand pushes the gel and the lead balls move accordingly.}

\begin{figure}[t]
    \centering
    \includegraphics[width=0.75\linewidth]{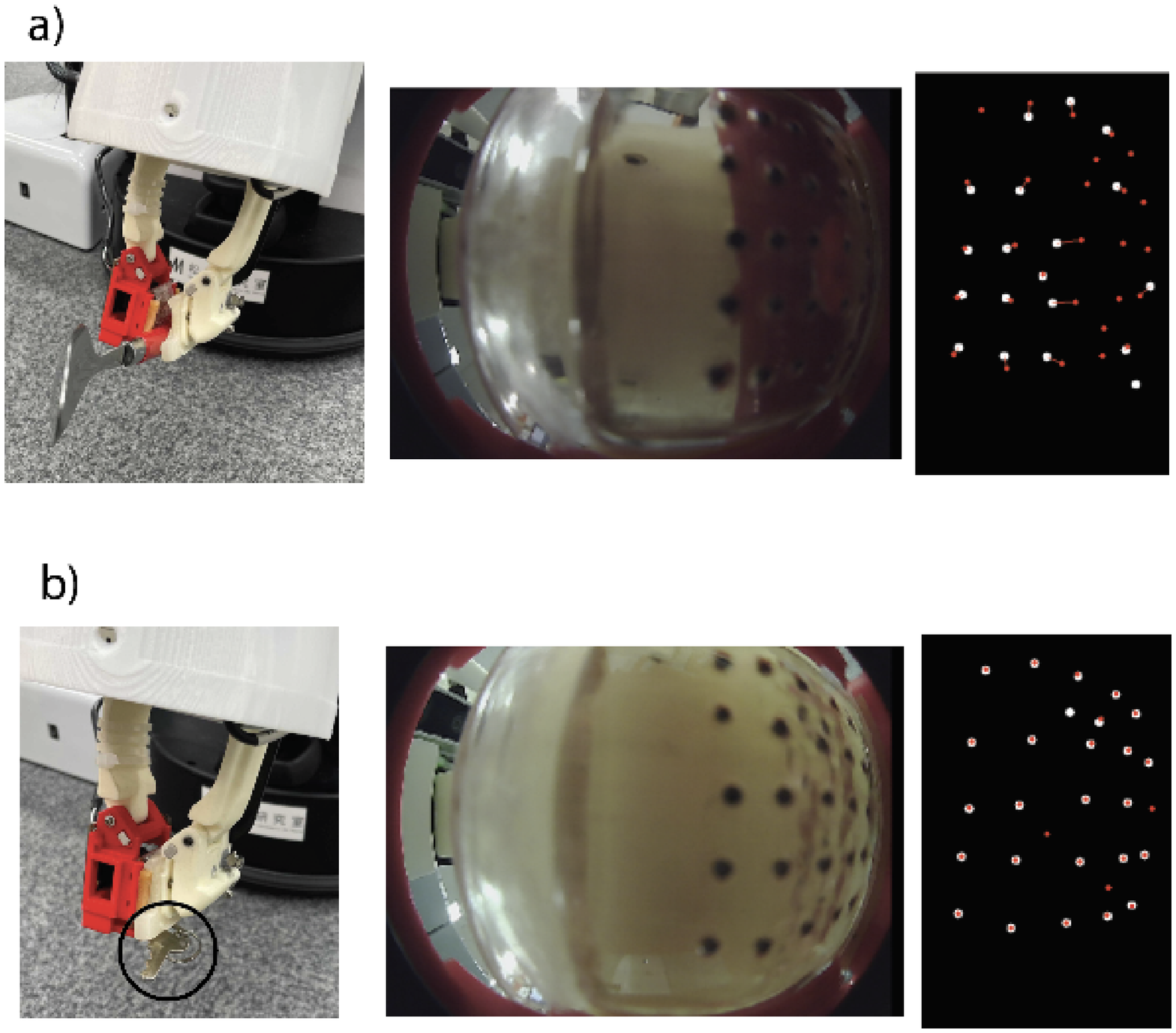}
    \caption{\red{Successful and failure cases of grasping.
    (a) Successful case. Spoon can be detected using the tactile sensor.
    (b) Failure case. The key hangs on between the edge parts of the tactile sensor module and the opposite gripper, which does not change the position of the markers over a threshold.}}
    \label{fig:grasp_case}
\end{figure}

Figure~\ref{fig:grasp_case}a is a successful example of grasp detection with a spoon.
In the right side of the image, the detected marker dots and red points are deviated and grasping can be detected.
This object could not be detected by the wrist sensor or the open-angle of the gripper.
Figure~\ref{fig:grasp_case}b represents a failure case.
The key was caught in the rubber fingernail~(Figure~\ref{fig:grasp_case}d) and almost no displacement occurred on the marker, as shown in Figure~\ref{fig:grasp_case}b .

\begin{figure}[t]
    \centering
    \includegraphics[width=0.75\linewidth]{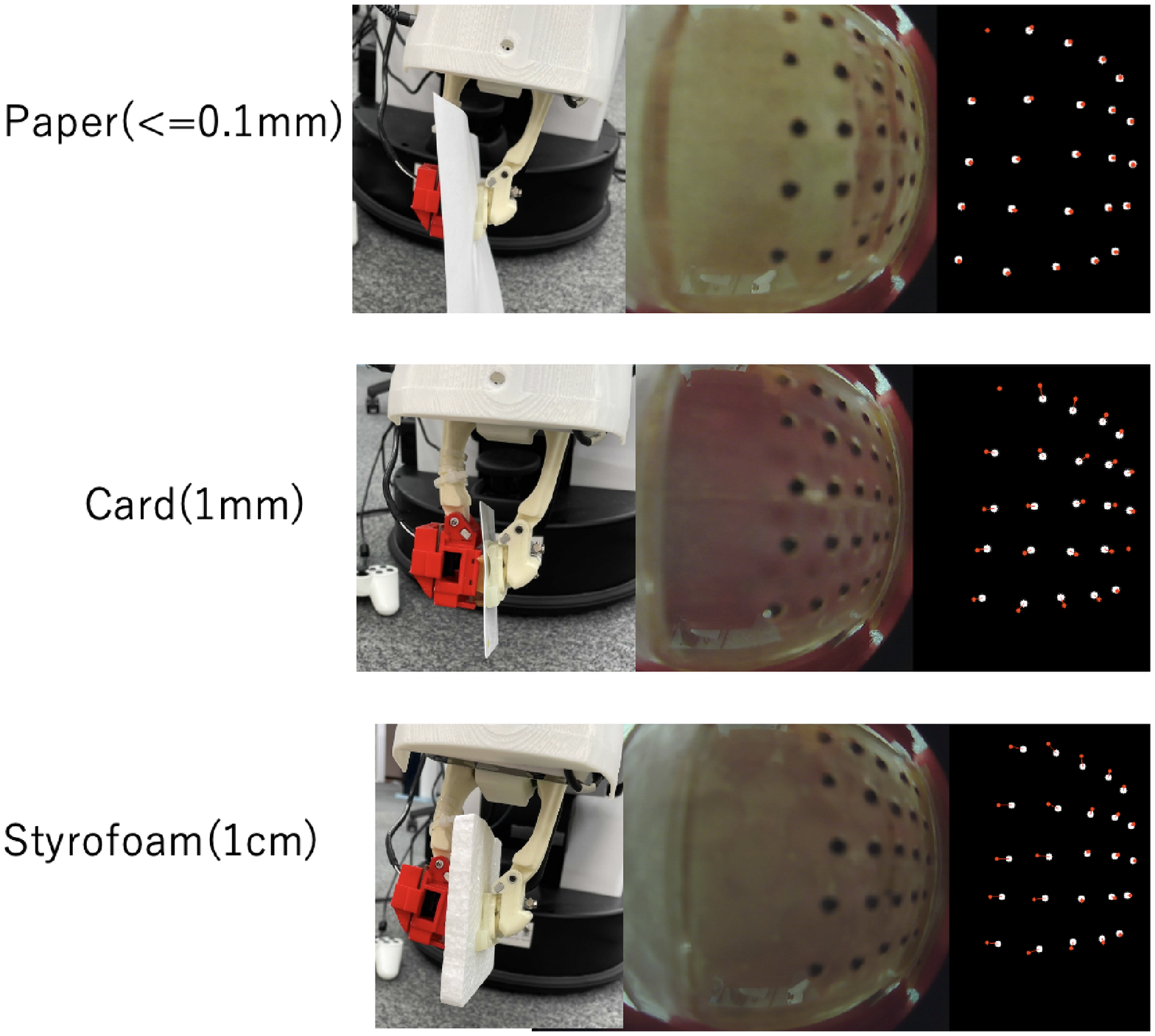}
    \caption{\red{Marker deviations when flat and objects are in hand; paper($\sim$0.1mm), a card($\sim$1mm), and a styrofoam board ($\sim$1cm).
    While paper is too thin to detect marker deviation with the proposed tactile sensor, it can detect the deviation of marker with considerably thin objects.}}
    \label{fig:flat_object}
\end{figure}

\red{
Figure~\ref{fig:flat_object} shows relationships between deviations of marker positions and thickness of flat objects in the hand.
Although the proposed tactile sensor is insensitive to a flat and extremely thin object like paper ($\sim$0.1mm) because the marker does not move a lot, it can detect grasp of considerably thin and flat object like a card ($\sim$1mm) or styrofoam board ($\sim$1cm) since the markers deviate to some extent.
}

\footnotetext{\url{https://docs.opencv.org/4.x/d7/d4d/tutorial_py_thresholding.html}}

\subsection{Navigation}
Moving from the current location to the destination while avoiding obstacles, a task termed navigation, is a basic operation for operating service robots at home.
For the navigation of household service robots, it is particularly important to accurately recognize objects that are difficult to recognize with sensors, such as small, transparent, or flexible objects.
Additionally, moving in response to the domestic environment, which changes constantly, is also important.

To navigate while avoiding collision with an object, it is necessary to have a positional relationship with the object.
Therefore, scan data and point cloud data that can directly acquire distance information are often used as sensors for navigation.
However, among various objects in a home, there are many objects that are difficult to recognize only from the shape information of the object (key, pen, clothes, etc.).

To avoid these objects, we use the following four data for obstacle information to be reflected in the costmap for path planning.

\begin{enumerate}
    \item Map created in advance (information on furniture and walls excluding objects). \label{nav1}
    \item 2D-LiDAR scan. \label{nav2}
    \item Point cloud filtered from the point cloud that can be taken with the depth camera. \label{nav3}
    \item Point cloud of the object recognized from image captured by the head RGB-D camera. \label{nav4}
\end{enumerate}

The information in (\ref{nav1}), (\ref{nav2}), and (\ref{nav3}) enables navigation by avoiding collisions with objects that can be recognized as shapes obtained from scan data and point clouds.

In (\ref{nav3}), the following filtering process is implemented.
The point cloud obtained from the depth camera is first downsampled and outliers removed.
Subsequently, the maximum height and minimum height of each voxel in the robot's base link coordinate are calculated, and if the value is below a threshold value, it is judged as the ground and removed.
And while the robot is carrying an object, it often determines its arm as an obstacle, so it removes the robot's own point cloud from link information and 3D model of the robot.

It is possible to avoid objects that cannot be seen from the 2D-LiDAR mounting position and objects that are indistinguishable from the ground by the point cloud height filtering process, by treating the object recognized by the camera in (\ref{nav4}) as an obstacle.

To reduce the impact of cumulative error in each estimation module, we implement position-based navigation relative to real-time recognized objects rather than absolute position-based navigation on pre-made maps.

Table \ref{tab:nav} lists the source of information on target position for navigation.
Except for the start position of each task, when moving to an object or furniture, the goal is calculated from the real-time recognition result.
This enables robust navigation to the cumulative error of self-position estimation by making heavy use of relative position-based navigation with recognition results, instead of absolute position-based navigation of pre-created maps.

\begin{table}[htbp]
    \centering
    \tbl{Recognition information used for the target position of the navigation.}{
    \begin{tabular}{c|c}
        Target Position & Recognition Information \\
        \hline\hline
        Object    & Object Recognition(\ref{sec:object_recognition})  \\
        Furniture & Sim-To-Real Image Segmentation(\ref{sec:sim2real_recognition}) \\
        Person    & Human Recognition(\ref{sec:human_recognition})  \\
        Task Start Point & Map
    \end{tabular}}
    \label{tab:nav}
\end{table}

However, it may occur several times while the robot is runnning that the route to the target  cannot be calculated due to an estimation error in the object recognition position.

If the robot cannot calculate the route to the goal for a certain period of time, the distance that the robot takes as a margin from the object used for path calculation is gradually narrowed from the point cloud of (\ref{nav4}).
At this time, instead of completely eliminating the cost of the narrowed part, we leave some cost, allowing the robot to move as far away from the object as possible after recalculating the route.

\begin{figure}
\begin{center}
    \subfigure[Third Person View]{
    \resizebox*{4cm}{!}{
        \includegraphics[width=0.3\linewidth]{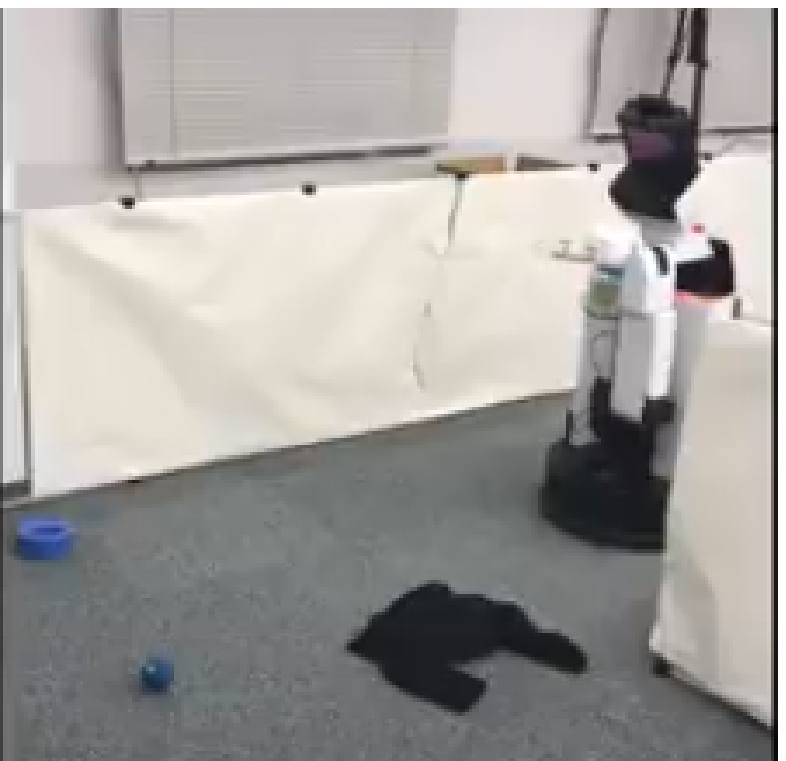}
        \label{subfig:3pv}
    }}
    \hspace{5pt}
    \subfigure[Recognition Result]{
    \resizebox*{4cm}{!}{
        \includegraphics[width=0.3\linewidth]{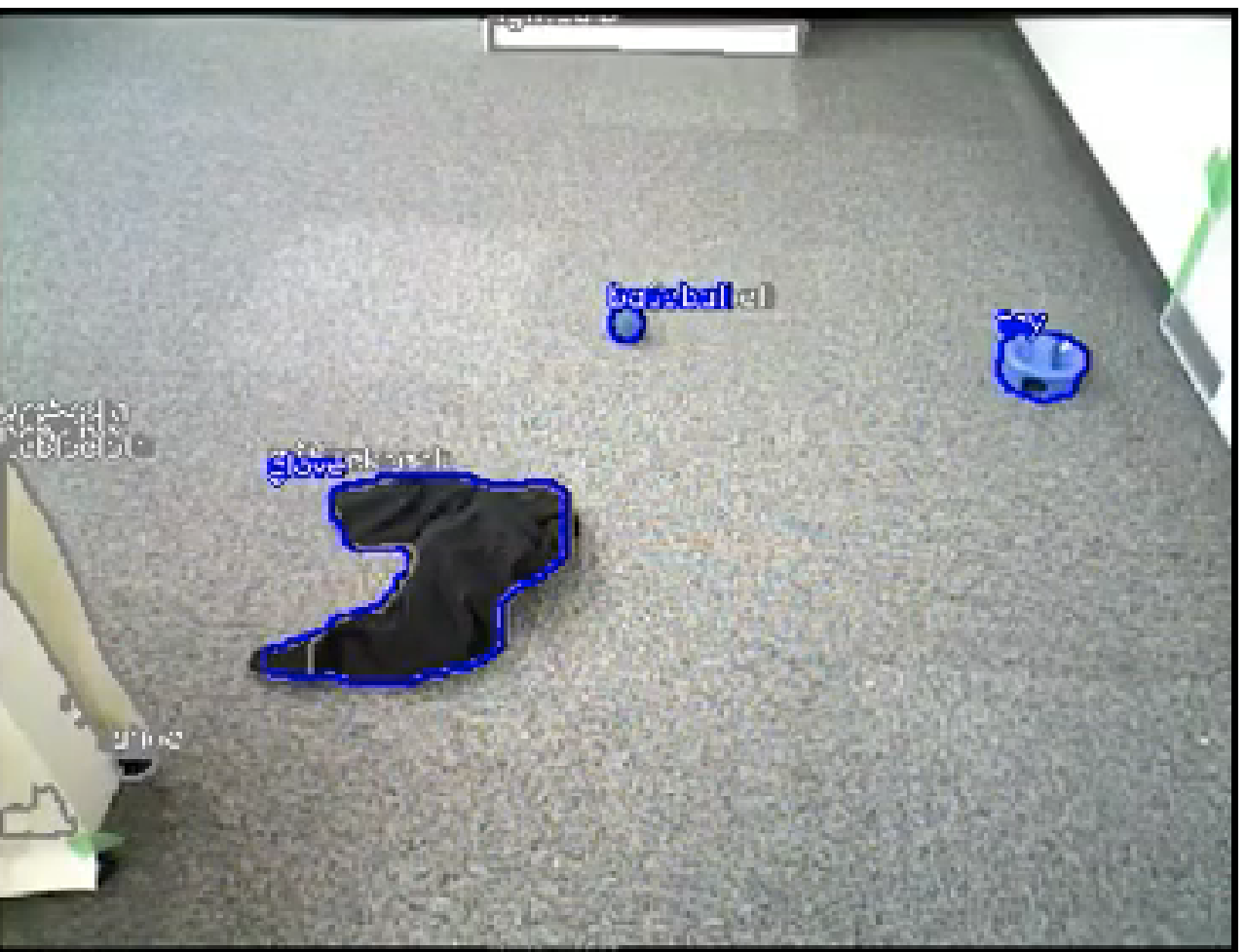}
        \label{subfig:task2a_detection}
    }}
    \hspace{5pt}
    \subfigure[Costmap reflecting recognition result]{
    \resizebox*{4cm}{!}{
        \includegraphics[width=0.3\linewidth]{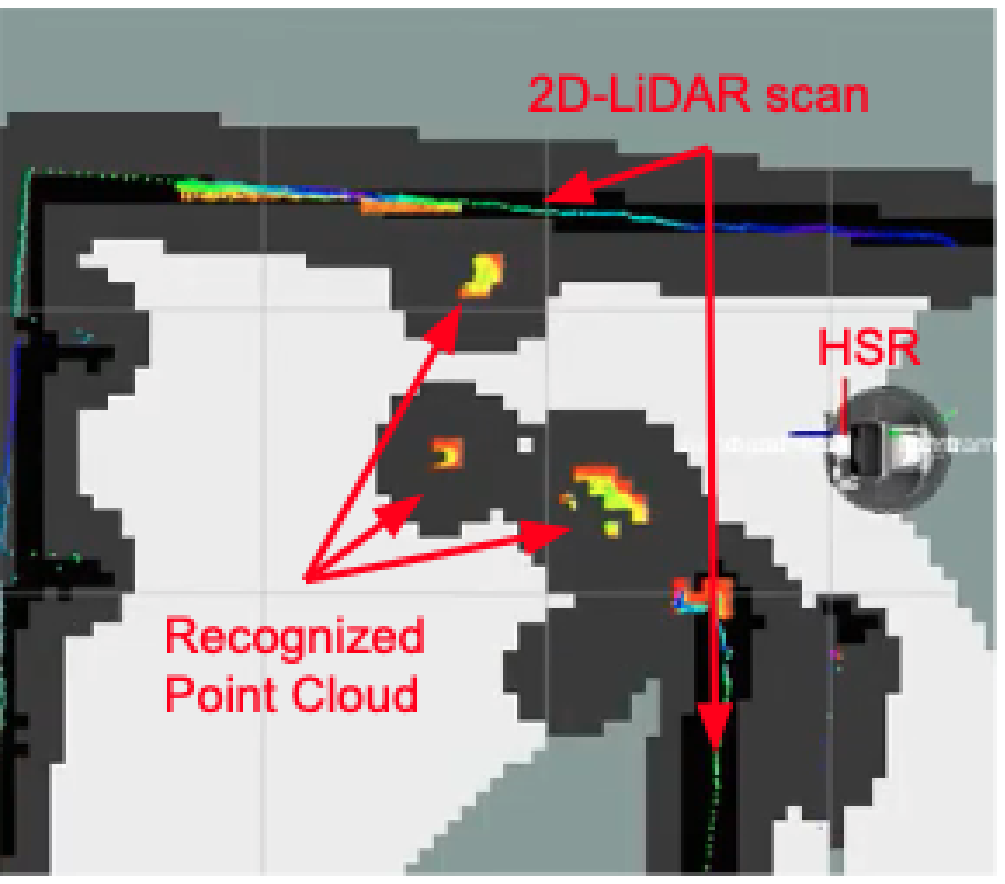}
        \label{subfig:task2a_costmap}
    }}
    \hspace{5pt}
    \caption{\red{An example of obstacle avoidance during navigation.
In an environment like (a), the robot recognizes obstacles in two ways.
The first one is to use 2D-LiDAR scan information to recognize obstacles above of a LiDAR sensor height e.g. large objects, furniture, and walls.
The second is to use RGB-D camera images to recognize low-height objects, such as clothes and balls that 2D-LiDAR cannot recognize.
The object surrounded by blue in (b) is the object recognized from the camera image.
Objects recognized in these two ways are reflected in the cost map in (c) used for path planning.}}
    \label{fig:nav}
\end{center}
\end{figure}

\red{
In the case of Figure~\ref{subfig:3pv}, small objects with a height that cannot be detected from 2D-LiDAR, namely, a pitcher lid, a small ball, and a T-shirt, are placed on the floor.
We detect these small objects from head-mounted RGB-D camera using object recognition modules described in the Section~\ref{sec:recognition} and cut out the corresponding point clouds of the objects as shown in Figure~\ref{subfig:task2a_detection}.
The costmap for navigation is updated using both LiDAR point cloud and detected object point clouds from head camera, and the path is planned as in Figure~\ref{subfig:task2a_costmap}.
Since this procedure of object detection, merging point clouds, and updating costmap and path plan is done in a closed-loop manner, the robot can avoid collisions dynamically.
}

\section{Integrated Experiment}
\label{sec:experiment}
\subsection{Preparation Time at the Site of the Competition}

At the WRC, each team brought their robot to the venue, and after four days of preparation, the competition was held over three days.
Two competition venues and three practice areas were set up, and the practice areas were made similar to the competition venue.

At the first step of preparation, our team created a map of the field using LiDAR mounted on HSR.
Subsequently, we captured pictures of objects at the venue and incorporated the data into the training model to tune the CLIP and other object recognition, primarily because the lighting at the venue was different from the environment in the laboratory, which may impair the recognition performance.

\red{In particular, in the competition, we used the object classifier fine-tuned with a dataset that we collected in the laboratory (748 images) and the competition venue (556 images).
We did not use the original YCB dataset, because the masking strategy is different from ours (explained in Section~\ref{subsubsec:object_detection}), which may cause deterioration of the recognition performance.
At the venue, we captured the images for the entire YCB objects by placing them on the floor of the arena for several times, and created a dataset of masked images using the object detection module (described in Section~\ref{subsubsec:object_detection}).
The class of corrected images was annotated manually, and the fully-connected layers of classifier were fine-tuned with the combined dataset.}

\subsection{Result}
Five teams competed at the real venue.
Two competition rooms had the same setup were set up at the venue, and the two teams started discussing at the same time, and the scores were used to determine the winner.
First, a round-robin tournament was held as a preliminary round, and then the top four teams competed in a tournament format.
Our team played six trials in total.
\red{Table~\ref{tab:wrs_result} lists the summarized scores changes~(see~Appendix~\ref{app:scoresheet} for the full score sheets).}

Our team got the best score in the semifinals (the fifth game).
In this case, we cleaned up 11 items in task1 and managed to complete task2 correctly.

For the task 1, the statistics for all six trials areas are noted in Table~\ref{tab:wrs_result_stats}.
\red{For object recognition~(category classification), the overall success rate across categories was 92\%} and ``Food'', ``Shape'', and ``Tool'' objects are perfectly categorized, while~\red{``task'' category had lower rate of 78\%.
In the competition, the some ``unknown'' objects outside YCB dataset~(not listed in the competition rule book) appeared, which were required to be stored in the place of either the corresponding category (announced in the trial) or ``unknown'' category.
We treated these unknown objects in the same way as the pre-announced objects in the rulebook by assuming that the recognition module generalized well even for the unknown objects, that is, we did not explicitly classify the ``known'' and ``unknown''.
The success rate of recognition of unknown objects was 50\% (2 out of 4).
}

The overall probability of successful grasping was 65\% in all six trials.
While ``Shape'' category had the highest success rate of 100\%, ``tool'' category had the lowest success rate of 45\%. The main reason was the difference of difficulty. ``Shape'' objects are mostly sphere shaped and easy to get good grasping position.
On the other hand, ``tool'' objects are relatively small and difficult items. 
In terms of task speed, the average time per object in task1, including failure and restart times, was 98.7 seconds. 
The average time per one successful object was 56.2 seconds.
It is quite slow compared to human cleanup speed but is deemed as a good result in the WRC2020 competition.

\begin{table}
    \centering
    \tbl{Competition results (6 trials)}{
    \begin{tabular}{c||c|c|c|c|c|c}
    Trial & 1st & 2nd & 3rd & 4th & 5th & 6th \\
    \hline \hline
    Task1 tidy up counts & 5 & 8 & 9 & 7 & 11 & 10 \\
    Task1 score & 145 & 185 & 240 & 145 & 320 & 250 \\
    Task2a success & True & False & True & False & True & True \\
    Task2b target & tomato soup can & sugar box & peer & apple & spam & chips can \\
    Task2 score & 390 & 200 & 390 & 100 & 410 & 150 \\
    Restart Robot & 1 & 0 & 1 & 1 & 0 & 2 \\
    Total Score & 535 & 385 & 630 & 245 & 730 & 400 \\
    \end{tabular}}
    \label{tab:wrs_result}
\end{table}

\begin{table}[]
    \centering
    \tbl{Success ratio of object \red{recognition} and tidy up in the competition. Since the boss item was placed at the same location in each trials, the \red{recognition} was not required.}{
    \begin{tabular}{c||c|c|c}
     Object Category & \red{Recognition} & Tidyup & \red{Both Recognition and Tidyup} \\
    \hline\hline
    Food(Listed in the rule book) & 1.0~\red{(11/11)} & 0.82~\red{(9/11)} & \red{0.72~(8/11)}   \\
    Kitchen(Listed in the rule book) & 0.88~\red{(14/16)} & 0.69~\red{(11/16)} & \red{0.5~(8/16)}  \\
    Shape(Listed in the rule book) & 1.0~\red{(11/11)} & 1.0~\red{(11/11)} & \red{1.0~(11/11)} \\
    Tool(Listed in the rule book) & 1.0~\red{(20/20)} & 0.45~\red{(9/20)} & \red{0.45~(9/20)} \\
    Task(Listed in the rule book) & 0.78~\red{(7/9)} & 0.56~\red{(5/9)} & \red{0.33~(3/9)} \\
    Unknown Objects~(\textbf{Not} Listed in the rule book) & \red{0.5~(3/6)} & \red{0.5~(3/6)} & \red{0.33~(2/6)} \\
    \red{Boss Item~(Shachihoko)} & \red{N/A} & \red{0.5~(2/4)} & \red{N/A} \\
    \hline
    Total & 0.92\red{(66/73)} & \red{0.65(50/77)} & \red{0.56(41/73)}  \\
    \end{tabular}}
    \label{tab:wrs_result_stats}
    
\end{table}

\section{Conclusion}
\label{sec:conclusion}
In the paper, we present the entire robot system developed for tidying up tasks that achieved the second prize in World Robot Challenge, a world-wide robot competition held in September 2021.
Our solution leveraged the data-driven approach for managing variations in home environments rather than directly pre-programming for edge-cases.
We demonstrate the generalization ability for deviations in the environment, and flexibility to update modules using corrected data during trials.
Moreover, we also demonstrate the system design for ensuring high throughput even if it contains modules deep neural networks, which requires high-computational loads.

\blue{
While the tidy-up task in WRC and our corresponding solution involves some essential aspects of service robots in home environments, for example, precise object recognition, object manipulation, and navigation, they are still insufficient for realizing generalist household robots.
For instance, safe manipulation and navigation in the environment with dynamic environment (e.g. moving objects and people), and social interaction between robots and human are the lacking aspects in the WRC task.
Further development and standardized benchmarks will bring generalist home robots into reality. 
}

As the future work, we plan to extend our robot system to accept massively corrected datasets by deployments~\cite{matsushima2021deploymentefficient} for adapting more diversified environments than those used in the WRC2020 competition, and establish a methodology for developments and operations of data-driven service robot systems.

\clearpage

\bibliographystyle{tADR}
\bibliography{0_0_arxiv_main}

\clearpage
\appendices

\section{Human Recognition}
\label{sec:human_recognition}

Service robots in household environments are often required to communicate and interact with humans to perform tasks.
In the task2, the goal of the task is to deliver the designated object to one of the persons requesting it by waving the hand.

We used Keypoint R-CNN~\citep{he2018mask} to detect people and their keypoints using input RGB images captured from a head-mounted camera. Simultaneously, we exclude any detected person that is outside the room by using depth images. 
Among the keypoints of the person detected in the room, we focused on the wrist and elbow to detect the person waving.
It was assumed that the target person raises the wrist higher than the other relative to the elbow.
When passing an object, the robot first moves to the waypoint position where it will not collide with the human, and then moves the end-effector in front of the target person.
The destination of the end-effector is set to the center coordinate of the target person's bounding box, which makes it possible to pass the object smoothly even when the target person is standing or sitting.
To achieve both high accuracy and time efficiency, we continue to count which person is waving until the difference in counts exceeds a threshold or a timeout period passes. We set the count difference to 3 and the timeout to 8 seconds, respectively.

\begin{figure}[t]
    \centering
    \includegraphics[width=0.9\linewidth]{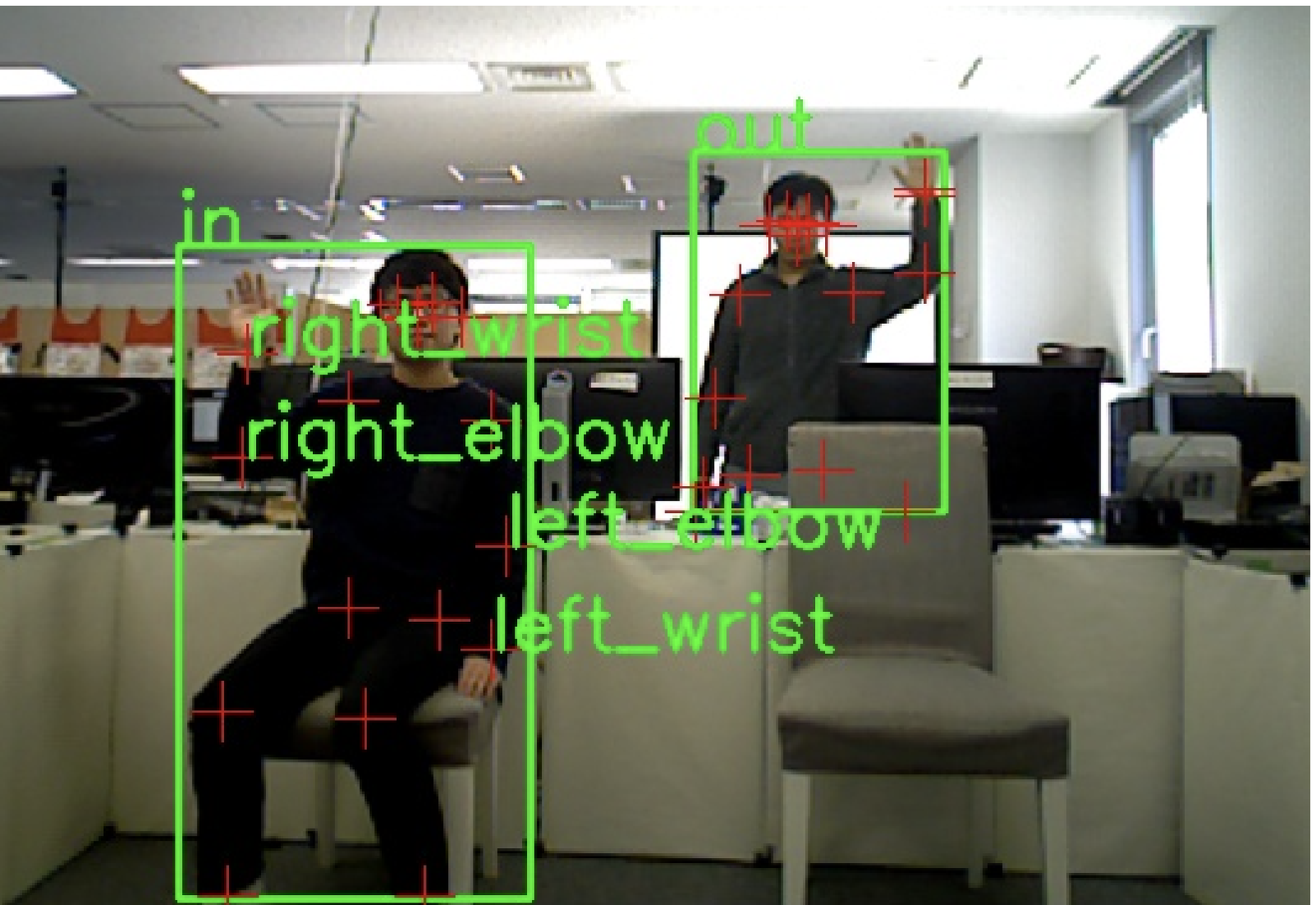}
    \caption{An example of person detection using Keypoint R-CNN.
Wrists and elbows are extracted from the obtained key points, and their positions are compared to determine if the person is waving a hand.
Additionally, using depth-based decision-making explained above makes it possible to detect a person who is lowering his hand from outside the field.}
    \label{fig:wave}
\end{figure}

We evaluated the method in this section from quantitative and qualitative perspectives.
First, as in Figure~\ref{fig:wave}, it is evident that the system could correctly determine that the right person is out of the room, showing that the system has succeeded in limiting the detection area as expected.

To analyze the accuracy, we randomly specified which target to wave and the pose of the target (standing or sitting), and the system achieved 100\% accuracy of target person detection in 10 trials with 0.43 seconds of inference time on average.

\section{\red{Other Grasp Pose Prediction Methods}}
\label{app:other_grasp}

\red{
Other than the grasp pose prediction methods described in the main text, we attempted several DNN-based approaches.
}

\paragraph*{\red{\textbf{Multi-Object Multi-Grasp}}}
\red{
Other methods use deep learning, such as a model inspired by object detection architectures~\cite{chu2018real}.
This method replaces the blue channel of an RGB image with depth and outputs oriented bounding boxes which represent 4 degrees of freedom (DoF) grasps with a parallel gripper. The model produced good predictions but only at a specific viewpoint, which is roughly top-down towards the object at about a meter away. This result can be explained by the fact that the model was trained on the Cornell Grasping Dataset~\cite{jiang2011efficient}, which contains RGB-D images shot from a single view.
}

\paragraph*{\red{\textbf{6-DoF GraspNet}}}
\red{
Another method we attempted was \red{6-Dof GraspNet}~\cite{mousavian20196}. This method can produce 6-DoF grasps, unlike the other methods described in this section, which are largely 4-DoF. The model generates many grasp candidates from a segmented point cloud of the target object. Despite high initial expectations, there were several major drawbacks. One was its computation time which was often consumed by its iterative grasp refinement method. Generating decent grasps for a single object took 3 to 10 seconds, which can be a considerable disadvantage in the competition. Another limitation was its brittleness; since the model expects a segmented point cloud, an inaccurate segmentation containing parts of the support surface or other objects significantly affected the reliability of the grasps. Third, the model does not consider collisions or kinematic feasibility, so we had to implement a post-processing function to cull such grasp candidates, which there were many. A later iteration of this model considers collisions~\cite{sundermeyer2021contact}, but we found that this was not always accurate and less so in non-tabletop situations.
}

\paragraph*{\red{\textbf{GGCNN}}}
\red{
Another principal method we attempted was GGCNN~\cite{morrison2020learning}. This model, like the grasp detector, also assumes a top-down viewpoint, but is more robust as it could deal with some variation in camera height. Though the work claimed that the model can update in real time, enabling faster calculation and possibly faster recovery from failed grasps, we could not reproduce this speed in the developed system. Regardless, we found this model to be inaccurate for many objects, as it often tried to grasp the edge of concave objects as if they had a graspable rim.
}

\newpage

\section{Raw Score Sheets in WRC Competition}
\label{app:scoresheet}
\red{The raw score sheets during the WRC competition are listed below. Superscripts on targets denote their category (food as $^f$, shape as $^s$, tool as $^{to}$, task as $^{ta}$ and kitchen as $^k$) and unknown objects are denoted by superscripts $^{unk}$.}

\begin{table}[h]
 \caption{\blue{1st trial (round-robin tournament \#1). The opponent is AISL-TUT whose score was 90~(90 for task1, and 0 for task2).}}
 \label{table:firsttrial}
 \centering
 \small
    \begin{tabular}{|cc|c:c:c|c:c:c|c|c|}
    \multicolumn{10}{c}{\textbf{Task 1}}\\
    \hline
    \multicolumn{2}{|c|}{} & \multicolumn{3}{|c|}{Penalties ($\times 0.5$)} & \multicolumn{5}{|c|}{Points} \\
    \hline
    \multicolumn{2}{|c|}{} & & & & \multicolumn{3}{|c|}{Correct ($+10$)} & False  & \\
    \cdashline{6-8}
   \# & Target & Restart & Drop & Hit & Delivery & Category & Orientation & ($\times 0.0$) & Total\\
   \hline
    1 & Softball $^s$ &
          &  &  &
        \checkmark & \checkmark & &
          & $20$ \\
    2 & Spoon $^k$ &
          &  &  &
          &  & &
          \checkmark & $0$ \\
    3 & Mini soccer ball $^s$ &
          & \checkmark &  &
        \checkmark & \checkmark & &
          & $10$ \\
    4 & Lemon $^f$ &
          &  &  &
        \checkmark & \checkmark & &
          & $20$ \\
    5 & Fork $^k$ &
          &  &  &
        \checkmark & \checkmark & \checkmark &
          & $30$ \\
    6 & Spoon $^k$ &
        \checkmark  &  &  &
        \checkmark & \checkmark & \checkmark &
          & $15$ \\
    
    \hline
    \multicolumn{9}{|l|}{Finishing task within the time limit ($+50$)} & $0$ \\
    \hline
    \multicolumn{9}{|l|}{Bonus challenge ($+50$/challenge)} &  \\
     & \multicolumn{8}{l|}{Boss character} & $0$ \\
     & \multicolumn{8}{l|}{Opening three drawers} & $50$ \\
     \hline
    \multicolumn{9}{r|}{SUBTOTAL (Task 1)} & $145$ \\
    \cline{10-10}
    
    \multicolumn{10}{c}{\textbf{Task 2}}  \\
    \hline
    \multicolumn{8}{|c|}{}  & Success & Points \\
    \hline
    \multicolumn{8}{|l|}{Task 2a} & & \\
     & \multicolumn{7}{l|}{Navigating to the goal without collision} & \checkmark & $100$ \\
    \hline
    \multicolumn{8}{|l|}{Task 2b} & & \\
     & \multicolumn{7}{l|}{Taking any food item in the shelf}  & \checkmark & $40$ \\
     & \multicolumn{7}{l|}{Taking the requested object}  & \checkmark & $60$ \\
     & \multicolumn{7}{l|}{Penalties ($25\%$/hit)}  & 0 hits & $-0$ \\
     \cdashline{2-10}
     & \multicolumn{7}{l|}{Delivering the object to a person}  & \checkmark & $40$ \\
     & \multicolumn{7}{l|}{Delivering the object to the requested person}  & \checkmark & $60$ \\
     & \multicolumn{7}{l|}{Penalties ($25\%$/hit)}  & 0 hits & $-0$ \\
    \hline
    \multicolumn{8}{|l|}{Finishing task within the time limit ($+50$)} & \checkmark & $50$ \\
    \multicolumn{8}{|l|}{If any time remaining, add 20 points per minute ($+20$/minute)} & $2$ min. & $40$ \\
    \hline
    \multicolumn{9}{r|}{SUBTOTAL (Task 2)} & $390$ \\
    \cline{10-10}
    \multicolumn{10}{c}{} \\
    \cline{10-10}
    \multicolumn{9}{r|}{\textbf{TOTAL (Task 1 + Task 2)}} & $\bm{535}$ \\
    \cline{10-10}
  \end{tabular}
\end{table}

\begin{table}[h]
 \caption{\blue{2nd trial (round-robin tournament \#2). The opponent is OIT-RITS whose score was 245~(30 for task1, and 215 for task2).}}
 \label{table:secondtrial}
 \centering
 \small
    \begin{tabular}{|cc|c:c:c|c:c:c|c|c|}
    \multicolumn{10}{c}{\textbf{Task 1}}\\
    \hline
    \multicolumn{2}{|c|}{} & \multicolumn{3}{|c|}{Penalties ($\times 0.5$)} & \multicolumn{5}{|c|}{Points} \\
    \hline
    \multicolumn{2}{|c|}{} & & & & \multicolumn{3}{|c|}{Correct ($+10$)} & False & \\
    \cdashline{6-8}
   \# & Target & Restart & Drop & Hit & Delivery & Category & Orientation & ($\times 0.0$) & Total\\
   \hline
    1 & Pitcher-Lid $^k$ &
          &  &  &
        \checkmark &  &  &
          & $10$ \\
    2 & Gelatin box $^f$ &
          &  &  &
        \checkmark & \checkmark & &
          & $20$ \\
    3 & Timer $^{ta}$ &
          &  &  &
        \checkmark & \checkmark & &
          & $20$ \\
    4 & 9 hole peg test $^{ta}$ &
          &  &  &
        \checkmark & \checkmark &  &
          & $20$ \\
    5 & Bowl $^k$ &
          &  &  &
        \checkmark & \checkmark &  &
          & $20$ \\
    6 & Large Marker $^{to}$ &
          &  &  &
        \checkmark & \checkmark & \checkmark &
          & $30$ \\
    7 & Rope $^s$ &
          &   & \checkmark &
        \checkmark & \checkmark &  &
          & $10$ \\
    8 & Pitcher-base $^k$ &
          &  & \checkmark &
        \checkmark &  &  &
          & $5$ \\
    9 & Small marker $^{to}$ &
          &  &  &
          &  &  &
        \checkmark  & $0$ \\
    10 & Watering can~$^{to, unk}$ &
          &  &  &
          &  &  &
        \checkmark  & $0$ \\
    11 & Small clamp $^{to}$ &
          &  &  &
         &  &  &
        \checkmark  & $0$ \\
    12 & Apple $^{f}$ &
          &  &  &
          &  &  &
        \checkmark  & $0$ \\
    
    \hline
    \multicolumn{9}{|l|}{Finishing task within the time limit ($+50$)} & $0$ \\
    \hline
    \multicolumn{9}{|l|}{Bonus challenge ($+50$/challenge)} &  \\
     & \multicolumn{8}{l|}{Boss character} & $0$ \\
     & \multicolumn{8}{l|}{Opening three drawers} & $50$ \\
     \hline
    \multicolumn{9}{r|}{SUBTOTAL (Task 1)} & $185$ \\
    \cline{10-10}
    
    \multicolumn{10}{c}{\textbf{Task 2}}  \\
    \hline
    \multicolumn{8}{|c|}{}  & Success & Points \\
    \hline
    \multicolumn{8}{|l|}{Task 2a} & & \\
     & \multicolumn{7}{l|}{Navigating to the goal without collision} &  & $0$ \\
    \hline
    \multicolumn{8}{|l|}{Task 2b} & & \\
     & \multicolumn{7}{l|}{Taking any food item in the shelf}  & \checkmark & $40$ \\
     & \multicolumn{7}{l|}{Taking the requested object}  & \checkmark & $60$ \\
     & \multicolumn{7}{l|}{Penalties ($25\%$/hit)}  & 0 hits & $-0$ \\
     \cdashline{2-10}
     & \multicolumn{7}{l|}{Delivering the object to a person}  & \checkmark & $40$ \\
     & \multicolumn{7}{l|}{Delivering the object to the requested person}  & \checkmark & $60$ \\
     & \multicolumn{7}{l|}{Penalties ($25\%$/hit)}  & 0 hits & $-0$ \\
    \hline
    \multicolumn{8}{|l|}{Finishing task within the time limit ($+50$)} &  & $0$ \\
    \multicolumn{8}{|l|}{If any time remaining, add 20 points per minute ($+20$/minute)} & $0$ min. & $0$ \\
    \hline
    \multicolumn{9}{r|}{SUBTOTAL (Task 2)} & $200$ \\
    \cline{10-10}
    \multicolumn{10}{c}{} \\
    \cline{10-10}
    \multicolumn{9}{r|}{\textbf{TOTAL (Task 1 + Task 2)}} & $\bm{385}$ \\
    \cline{10-10}
  \end{tabular}
\end{table}

\begin{table}[h]
 \caption{\blue{3rd trial (round-robin tournament \#3). The opponent is HIBIKINO-Musashi@Home whose score was 652~(222 for task1, and 430 for task2).}}
 \label{table:thirdtrial}
 \centering
 \small
    \begin{tabular}{|cc|c:c:c|c:c:c|c|c|}
    \multicolumn{10}{c}{\textbf{Task 1}}\\
    \hline
    \multicolumn{2}{|c|}{} & \multicolumn{3}{|c|}{Penalties ($\times 0.5$)} & \multicolumn{5}{|c|}{Points} \\
    \hline
    \multicolumn{2}{|c|}{} & & & & \multicolumn{3}{|c|}{Correct ($+10$)} & False & \\
    \cdashline{6-8}
   \# & Target & Restart & Drop & Hit & Delivery & Category & Orientation & ($\times 0.0$) & Total\\
   \hline
    1 & Pitcher-Lid $^{k}$ &
          &  &  &
        \checkmark &  &  &
          & $10$ \\
    2 & Cucumber~$^{f, unk}$ &
          &  &  &
        \checkmark & \checkmark & &
          & $20$ \\
    3 & Large Marker $^{to}$ &
          &  &  &
        \checkmark & \checkmark & \checkmark &
          & $30$ \\
    4 & Spam $^{f}$&
          &  &  &
        \checkmark & \checkmark &  &
          & $20$ \\
    5 & Large clamp $^{to}$ &
          &  &  &
        \checkmark & \checkmark &  &
          & $20$ \\
    6 & Cups $^{k}$ &
          &  &  &
        \checkmark & \checkmark &  &
          & $20$ \\
    7 & Small marker $^{to}$ &
          &  &  &
          &  &  &
        \checkmark  & $0$ \\
    8 & Rubik's cube $^{ta}$ &
          &  &  &
          &  &  &
        \checkmark  & $0$ \\
    9 & Bowl $^{k}$ &
          & \checkmark  & \checkmark &
        \checkmark & \checkmark &  &
          & $5$ \\
    10 & Pitcher-base $^{k}$ &
          &  & \checkmark &
        \checkmark & \checkmark &  &
          & $10$ \\
    11 & 9 peg hole test $^{ta}$ &
          &  &  &
          &  &  &
        \checkmark & $0$ \\
    12 & Boss &
         \checkmark & \checkmark &  &
        \checkmark & \checkmark &  &
          & $5$ \\
    
    \hline
    \multicolumn{9}{|l|}{Finishing task within the time limit ($+50$)} & $0$ \\
    \hline
    \multicolumn{9}{|l|}{Bonus challenge ($+50$/challenge)} &  \\
     & \multicolumn{8}{l|}{Boss character} & $50$ \\
     & \multicolumn{8}{l|}{Opening three drawers} & $50$ \\
     \hline
    \multicolumn{9}{r|}{SUBTOTAL (Task 1)} & $240$ \\
    \cline{10-10}
    
    \multicolumn{10}{c}{\textbf{Task 2}}  \\
    \hline
    \multicolumn{8}{|c|}{}  & Success & Points \\
    \hline
    \multicolumn{8}{|l|}{Task 2a} & & \\
     & \multicolumn{7}{l|}{Navigating to the goal without collision} & \checkmark & $100$ \\
    \hline
    \multicolumn{8}{|l|}{Task 2b} & & \\
     & \multicolumn{7}{l|}{Taking any food item in the shelf}  & \checkmark & $40$ \\
     & \multicolumn{7}{l|}{Taking the requested object}  & \checkmark & $60$ \\
     & \multicolumn{7}{l|}{Penalties ($25\%$/hit)}  & 0 hits & $-0$ \\
     \cdashline{2-10}
     & \multicolumn{7}{l|}{Delivering the object to a person}  & \checkmark & $40$ \\
     & \multicolumn{7}{l|}{Delivering the object to the requested person}  & \checkmark & $60$ \\
     & \multicolumn{7}{l|}{Penalties ($25\%$/hit)}  & 0 hits & $-0$ \\
    \hline
    \multicolumn{8}{|l|}{Finishing task within the time limit ($+50$)} & \checkmark & $50$ \\
    \multicolumn{8}{|l|}{If any time remaining, add 20 points per minute ($+20$/minute)} & $2$ min. & $40$ \\
    \hline
    \multicolumn{9}{r|}{SUBTOTAL (Task 2)} & $390$ \\
    \cline{10-10}
    \multicolumn{10}{c}{} \\
    \cline{10-10}
    \multicolumn{9}{r|}{\textbf{TOTAL (Task 1 + Task 2)}} & $\bm{630}$ \\
    \cline{10-10}
  \end{tabular}
\end{table}

\begin{table}[h]
 \caption{\blue{4th trial (round-robin tournament \#4). The opponent is eR@sers whose score was 85~(85 for task1, and 0 for task2).}}
 \label{table:forthtrial}
 \centering
 \small
    \begin{tabular}{|cc|c:c:c|c:c:c|c|c|}
    \multicolumn{10}{c}{\textbf{Task 1}}\\
    \hline
    \multicolumn{2}{|c|}{} & \multicolumn{3}{|c|}{Penalties ($\times 0.5$)} & \multicolumn{5}{|c|}{Points} \\
    \hline
    \multicolumn{2}{|c|}{} & & & & \multicolumn{3}{|c|}{Correct ($+10$)} & False & \\
    \cdashline{6-8}
   \# & Target & Restart & Drop & Hit & Delivery & Category & Orientation & ($\times 0.0$) & Total\\
   \hline
    1 & Metal mug $^{k}$ &
          &  &  &
        \checkmark & \checkmark &  &
          & $20$ \\
    2 & Cookie box~$^{f, unk}$ &
          &  &  &
        \checkmark & \checkmark & &
          & $20$ \\
    3 & Small marker$^{to}$ &
          &  &  &
          &  &  &
        \checkmark  & $0$ \\
    4 & Strawberry $^{f}$&
          & \checkmark &  &
        \checkmark & \checkmark &  &
          & $10$ \\
    5 & Timer $^{ta}$&
          &  & \checkmark &
        \checkmark &  &  &
          & $5$ \\
    5 & Small marker $^{to}$&
          &  &  &
          &  &  &
        \checkmark & $0$ \\
    6 & T-shirt $^{ta}$&
          &  &  &
          &  &  &
        \checkmark  & $0$ \\
    7 & Spatula $^{k}$&
          &  &  &
          &  &  &
        \checkmark  & $0$ \\
    8 & Nuts $^{to}$&
          &  &  &
          &  &  &
        \checkmark  & $0$ \\
    9 & Clamp $^{to}$&
          & \checkmark &  &
        \checkmark & \checkmark &  &
          & $10$ \\
    10 & Form brick $^{s}$&
          &  &  &
        \checkmark & \checkmark &  &
          & $20$ \\
    11 & Spoon $^{k}$&
          &  &  &
          &  &  &
        \checkmark  & $0$ \\
    12 & Large clamp $^{to}$&
          & \checkmark &  &
        \checkmark & \checkmark &  &
          & $10$ \\
    13 & 9 hole peg test $^{ta}$&
        \checkmark  & \checkmark &  &
          &  &  &
        \checkmark  & $0$ \\
    
    \hline
    \multicolumn{9}{|l|}{Finishing task within the time limit ($+50$)} & $0$ \\
    \hline
    \multicolumn{9}{|l|}{Bonus challenge ($+50$/challenge)} &  \\
     & \multicolumn{8}{l|}{Boss character} & $0$ \\
     & \multicolumn{8}{l|}{Opening three drawers} & $50$ \\
     \hline
    \multicolumn{9}{r|}{SUBTOTAL (Task 1)} & $145$ \\
    \cline{10-10}
    
    \multicolumn{10}{c}{\textbf{Task 2}}  \\
    \hline
    \multicolumn{8}{|c|}{}  & Success & Points \\
    \hline
    \multicolumn{8}{|l|}{Task 2a} & & \\
     & \multicolumn{7}{l|}{Navigating to the goal without collision} &  & $0$ \\
    \hline
    \multicolumn{8}{|l|}{Task 2b} & & \\
     & \multicolumn{7}{l|}{Taking any food item in the shelf}  & \checkmark & $40$ \\
     & \multicolumn{7}{l|}{Taking the requested object}  & \checkmark & $60$ \\
     & \multicolumn{7}{l|}{Penalties ($25\%$/hit)}  & 0 hits & $-0$ \\
     \cdashline{2-10}
     & \multicolumn{7}{l|}{Delivering the object to a person}  &  & $0$ \\
     & \multicolumn{7}{l|}{Delivering the object to the requested person}  &  & $0$ \\
     & \multicolumn{7}{l|}{Penalties ($25\%$/hit)}  & 0 hits & $-0$ \\
    \hline
    \multicolumn{8}{|l|}{Finishing task within the time limit ($+50$)} &  & $0$ \\
    \multicolumn{8}{|l|}{If any time remaining, add 20 points per minute ($+20$/minute)} & $2$ min. & $0$ \\
    \hline
    \multicolumn{9}{r|}{SUBTOTAL (Task 2)} & $100$ \\
    \cline{10-10}
    \multicolumn{10}{c}{} \\
    \cline{10-10}
    \multicolumn{9}{r|}{\textbf{TOTAL (Task 1 + Task 2)}} & $\bm{245}$ \\
    \cline{10-10}
  \end{tabular}
\end{table}

\begin{table}[h]
 \caption{\blue{5th trial (semi-final). The opponent is eR@sers whose score was 342~(102 for task1, and 240 for task2).}}
 \label{table:fifthtrial}
 \centering
 \small
    \begin{tabular}{|cc|c:c:c|c:c:c|c|c|}
    \multicolumn{10}{c}{\textbf{Task 1}}\\
    \hline
    \multicolumn{2}{|c|}{} & \multicolumn{3}{|c|}{Penalties ($\times 0.5$)} & \multicolumn{5}{|c|}{Points} \\
    \hline
    \multicolumn{2}{|c|}{} & & & & \multicolumn{3}{|c|}{Correct ($+10$)} & False & \\
    \cdashline{6-8}
   \# & Target & Restart & Drop & Hit & Delivery & Category & Orientation & ($\times 0.0$) & Total\\
   \hline
    1 & Chips can $^{f}$&
          &  &  &
        \checkmark & \checkmark &  &
          & $20$ \\
    2 & Peach $^{f}$&
          &  &  &
        \checkmark & \checkmark & &
          & $20$ \\
    3 & Spoon $^{k}$&
          &  &  &
        \checkmark & \checkmark & \checkmark &
          & $30$ \\
    4 & O'ball~$^{s, unk}$&
          &  &  &
          &  &  &
        \checkmark  & $0$ \\
    5 & Muscat~$^{f, unk}$&
          &  &  &
        \checkmark &  &  &
          & $10$ \\
    6 & Racket ball $^{s}$ &
          &  &  &
        \checkmark & \checkmark &  &
          & $20$ \\
    7 & Small marker $^{to}$&
          &  &  &
          &  &  &
        \checkmark  & $0$ \\
    8 & Spatula $^{k}$&
          &  &  &
         &  &  &
        \checkmark  & $0$ \\
    9 & Small marker $^{to}$&
          &  &  &
        \checkmark & \checkmark & \checkmark &
          & $30$ \\
    10 & Wine glass $^{k}$&
          &   &  &
        \checkmark & \checkmark &  &
          & $20$ \\
    11 & Foam brick $^{s}$&
          &  &  &
        \checkmark & \checkmark &  &
          & $20$ \\
    12 & Boss &
          &  & \checkmark &
        \checkmark & \checkmark &  &
          & $10$ \\
    13 & Rope $^{s}$&
          &  &  &
        \checkmark & \checkmark &  &
          & $20$ \\
    14 & Tomato soup can $^{f}$&
          &  &  &
        \checkmark & \checkmark &  &
          & $20$ \\
    
    \hline
    \multicolumn{9}{|l|}{Finishing task within the time limit ($+50$)} & $0$ \\
    \hline
    \multicolumn{9}{|l|}{Bonus challenge ($+50$/challenge)} &  \\
     & \multicolumn{8}{l|}{Boss character} & $50$ \\
     & \multicolumn{8}{l|}{Opening three drawers} & $50$ \\
     \hline
    \multicolumn{9}{r|}{SUBTOTAL (Task 1)} & $320$ \\
    \cline{10-10}
    
    \multicolumn{10}{c}{\textbf{Task 2}}  \\
    \hline
    \multicolumn{8}{|c|}{}  & Success & Points \\
    \hline
    \multicolumn{8}{|l|}{Task 2a} & & \\
     & \multicolumn{7}{l|}{Navigating to the goal without collision} & \checkmark & $100$ \\
    \hline
    \multicolumn{8}{|l|}{Task 2b} & & \\
     & \multicolumn{7}{l|}{Taking any food item in the shelf}  & \checkmark & $40$ \\
     & \multicolumn{7}{l|}{Taking the requested object}  & \checkmark & $60$ \\
     & \multicolumn{7}{l|}{Penalties ($25\%$/hit)}  & 0 hits & $-0$ \\
     \cdashline{2-10}
     & \multicolumn{7}{l|}{Delivering the object to a person}  & \checkmark & $40$ \\
     & \multicolumn{7}{l|}{Delivering the object to the requested person}  & \checkmark & $60$ \\
     & \multicolumn{7}{l|}{Penalties ($25\%$/hit)}  & 0 hits & $-0$ \\
    \hline
    \multicolumn{8}{|l|}{Finishing task within the time limit ($+50$)} & \checkmark & $50$ \\
    \multicolumn{8}{|l|}{If any time remaining, add 20 points per minute ($+20$/minute)} & $3$ min. & $60$ \\
    \hline
    \multicolumn{9}{r|}{SUBTOTAL (Task 2)} & $410$ \\
    \cline{10-10}
    \multicolumn{10}{c}{} \\
    \cline{10-10}
    \multicolumn{9}{r|}{\textbf{TOTAL (Task 1 + Task 2)}} & $\bm{730}$ \\
    \cline{10-10}
  \end{tabular}
\end{table}

\begin{table}[h]
 \caption{\blue{6th trial (final). The opponent is Hibikino-Musashi@Home whose score was 800~(390 for task1, and 410 for task2).}}
 \label{table:sixthtrial}
 \centering
 \small
    \begin{tabular}{|cc|c:c:c|c:c:c|c|c|}
    \multicolumn{10}{c}{\textbf{Task 1}}\\
    \hline
    \multicolumn{2}{|c|}{} & \multicolumn{3}{|c|}{Penalties ($\times 0.5$)} & \multicolumn{5}{|c|}{Points} \\
    \hline
    \multicolumn{2}{|c|}{} & & & & \multicolumn{3}{|c|}{Correct ($+10$)} & False & \\
    \cdashline{6-8}
   \# & Target & Restart & Drop & Hit & Delivery & Category & Orientation & ($\times 0.0$) & Total\\
   \hline
    1 & Rubik's cube $^{ta}$&
          &  &  &
        \checkmark & \checkmark &  &
          & $20$ \\
    2 & Large clamp $^{to}$&
          &  &  &
        \checkmark & \checkmark & &
          & $20$ \\
    3 & Orange $^{f}$&
          &  &  &
         &  &  &
        \checkmark  & $0$ \\
    4 & Cracker box $^{f}$&
          & \checkmark &  &
          &  &  &
        \checkmark  & $0$ \\
    5 & Orange $^{f}$&
          & \checkmark &  &
        \checkmark & \checkmark &  &
          & $10$ \\
    6 & Small marker $^{to}$&
          &  &  &
        \checkmark & \checkmark & \checkmark &
          & $30$ \\
    7 & Cups $^{k}$&
          &  &  &
        \checkmark & \checkmark &  &
          & $20$ \\
    8 & Tennis ball $^{s}$&
          &  &  &
        \checkmark & \checkmark &  &
          & $20$ \\
    9 & Banana $^{f}$&
          &   &  &
        \checkmark & \checkmark &  &
          & $20$ \\
    10 & Mini soccer ball $^{s}$&
          &  &  &
        \checkmark & \checkmark &  &
          & $20$ \\
    11 & Large marker $^{to}$&
          &  &  &
        \checkmark & \checkmark & \checkmark &
          & $30$ \\
    12 & Wood block $^{s}$&
          &  &  &
        \checkmark &  &  &
          & $10$ \\
    13 & Boss &
        \checkmark  &  &  &
          &  &  &
        \checkmark  & $0$ \\
    
    \hline
    \multicolumn{9}{|l|}{Finishing task within the time limit ($+50$)} & $0$ \\
    \hline
    \multicolumn{9}{|l|}{Bonus challenge ($+50$/challenge)} &  \\
     & \multicolumn{8}{l|}{Boss character} & $0$ \\
     & \multicolumn{8}{l|}{Opening three drawers} & $50$ \\
     \hline
    \multicolumn{9}{r|}{SUBTOTAL (Task 1)} & $250$ \\
    \cline{10-10}
    
    \multicolumn{10}{c}{\textbf{Task 2}}  \\
    \hline
    \multicolumn{8}{|c|}{}  & Success & Points \\
    \hline
    \multicolumn{8}{|l|}{Task 2a} & & \\
     & \multicolumn{7}{l|}{Navigating to the goal without collision} & \checkmark & $100$ \\
    \hline
    \multicolumn{8}{|l|}{Task 2b} & & \\
     & \multicolumn{7}{l|}{Taking any food item in the shelf}  & \checkmark & $40$ \\
     & \multicolumn{7}{l|}{Taking the requested object}  & \checkmark & $60$ \\
     & \multicolumn{7}{l|}{Penalties ($25\%$/hit)}  & 2 hits & $-50$ \\
     \cdashline{2-10}
     & \multicolumn{7}{l|}{Delivering the object to a person}  &  & $0$ \\
     & \multicolumn{7}{l|}{Delivering the object to the requested person}  &  & $0$ \\
     & \multicolumn{7}{l|}{Penalties ($25\%$/hit)}  & 0 hits & $-0$ \\
    \hline
    \multicolumn{8}{|l|}{Finishing task within the time limit ($+50$)} &  & $0$ \\
    \multicolumn{8}{|l|}{If any time remaining, add 20 points per minute ($+20$/minute)} & $0$ min. & $0$ \\
    \hline
    \multicolumn{9}{r|}{SUBTOTAL (Task 2)} & $150$ \\
    \cline{10-10}
    \multicolumn{10}{c}{} \\
    \cline{10-10}
    \multicolumn{9}{r|}{\textbf{TOTAL (Task 1 + Task 2)}} & $\bm{400}$ \\
    \cline{10-10}
  \end{tabular}
\end{table}

\label{lastpage}

\end{document}